\documentclass[11pt]{article} 
\usepackage{url}
\usepackage{smile}
\usepackage{graphicx} 
\usepackage{epstopdf}
\usepackage{wrapfig}
\usepackage[colorlinks, linkcolor=blue, anchorcolor=blue, citecolor=blue]{hyperref}
\usepackage[margin=1in]{geometry}
\usepackage[normalem]{ulem}
\usepackage[export]{adjustbox} 
\usepackage{mathtools, cuted}
\usepackage{enumerate}
\usepackage{enumitem}
\usepackage{microtype}
\usepackage{graphicx}
\usepackage{booktabs}
\usepackage{multirow} 
\usepackage{subcaption}
\usepackage{amsmath}
\usepackage{amssymb}
\usepackage{mathtools}
\usepackage{amsthm}
\usepackage{algorithm}
\usepackage{algorithmic}
\usepackage{wrapfig}

\usepackage[margin=1in]{geometry}
\usepackage[normalem]{ulem}
\usepackage[export]{adjustbox}
\usepackage{mathtools, cuted}
\usepackage{natbib}
\usepackage{enumerate}
\usepackage{enumitem}

\usepackage{kpfonts}
\DeclareMathAlphabet{\mathsf}{OT1}{cmss}{m}{n}

\SetMathAlphabet{\mathsf}{bold}{OT1}{cmss}{bx}{n}

\newtheorem*{theorem*}{Theorem}

\newcommand{\bTheta}{\bm{\theta}}
\newcommand{\bThetat}{\bTheta^{(t)}}
\newcommand{\bThetaj}{\bTheta_{j,-j}}

\newcommand{\bThetaGj}{\bTheta_{\mathcal{G}_{j}}}

\newcommand{\GG}{\mathcal{G}}
\newcommand{\Gj}{\GG_{j}}
\newcommand{\I}{I}
\newcommand{\It}{\I^{(t)}}

\newcommand{\Itj}{\It_{j}}

\newcommand{\Ibar}{\overline{\I}}
\newcommand{\Ibarj}{\Ibar_{j}}
\newcommand{\Ibart}{\Ibar^{(t)}}

\newcommand{\Ibartj}{\Ibart_{j}}

\newcommand{\betaI}{\beta_{1}}
\newcommand{\betaU}{\beta_{2}}
\newcommand{\U}{U}
\newcommand{\Ut}{\U^{(t)}}
\newcommand{\Utj}{\Ut_{j}}
\newcommand{\Ubar}{\overline{\U}}
\newcommand{\Ubarj}{\Ubar_{j}}
\newcommand{\Ubart}{\Ubar^{(t)}}
\newcommand{\Ubartj}{\Ubart_{j}}
\newcommand{\DD}{\mathcal{D}} %

\newcommand{\Proj}{\mathcal{T}}
\newcommand{\rr}{r}
\newcommand{\rt}{\rr^{(t)}}

\newcommand{\Sc}{S}
\newcommand{\Sct}{\Sc^{(t)}}

\newcommand{\OurAlg}{PLATON}

\title{\bf PLATON: Pruning Large Transformer Models with Upper Confidence Bound of Weight Importance\footnote{Published as a conference paper in ICML 2022.}}

\author{Qingru Zhang, Simiao Zuo, Chen Liang, Alexander Bukharin \\ Pengcheng He, Weizhu Chen, Tuo Zhao \footnote{Zhang, Zuo, Liang, Bukharin and Zhao are affiliated with Georgia Tech. He and Chen are affiliated with Microsoft Azure. Correspondence to \url{qingru.zhang@gatech.edu} and \url{tourzhao@gatech.edu}.}}

\newcommand{\commentout}[1]{}

\begin{document}

\maketitle

\begin{abstract}
	Large Transformer-based models have exhibited superior performance in various natural language processing and computer vision tasks. However, these models contain enormous amounts of parameters, which restrict their deployment to real-world applications. To reduce the model size, researchers prune these models based on the weights' importance scores. However, such scores are usually estimated on mini-batches during training, which incurs large variability/uncertainty due to mini-batch sampling and complicated training dynamics. As a result, some crucial weights could be pruned by commonly used pruning methods because of such uncertainty, which makes training unstable and hurts generalization. To resolve this issue, we propose PLATON, 
	which captures the uncertainty of importance scores
	by upper confidence bound (UCB) of importance estimation.  
	In particular, for the weights with low importance scores but high uncertainty, {\OurAlg} tends to retain them and explores their capacity. We conduct extensive experiments with several Transformer-based models on natural language understanding, question answering and image classification to validate the effectiveness of {\OurAlg}. Results demonstrate that {\OurAlg} manifests notable improvement under different sparsity levels. 
	Our code is publicly available at \url{https://github.com/QingruZhang/PLATON}.
\end{abstract}


\section{Introduction}\label{sec:introduction}
Large Transformer-based models have exhibited superior performance in various tasks, such as natural language processing \citep{devlin2018bert,liu2019roberta,he2021deberta,radford2019language,brown2020language} and computer vision \citep{dosovitskiy2020image}.
These models advance start-of-the-art results in natural language understanding (e.g, GLUE, \citealt{wang2018glue}), question answering (e.g. SQuAD, \citealt{squad1}) and image classification (e.g., pImageNet, \citealt{deng2009imagenet}).
However, they typically incur massive memory footprint, e.g., BERT models \citep{devlin2018bert} contain up to 345 million parameters; Vision transformer models (ViT, \citealt{dosovitskiy2020image}) consist of up to 300 million parameters and GPT-3 models \citep{brown2020language} comprise up to 175 billion parameters.
Consequently, it becomes very challenging and expensive to deploy these models to real-world applications due to memory and energy consumption requirements. Moreover, these models' significant inference latency raises a barrier for their practicality, especially for high throughput environments such as e-commerce search platforms. 

Confronting the abovementioned challenges, pruning methods are widely applied
at only a small expense of model performance. These methods prune the parameters based on certain importance scores to significantly reduce model sizes \citep{han2015learning,han2016eie,zhu2017prune}. Popular importance scores are based on the parameters' magnitude  \citep{han2015learning,han2015deep,paganini2020iterative} or sensitivity \citep{molchanov2016pruning,molchanov2019importance,ding2019global,sanh2020movement,liang2021super}. Parameters with low importance scores are considered redundant and expected to be pruned.

Existing pruning methods mainly fall into two categories: \emph{one-shot pruning} \citep{lee2018snip,frankle2018lottery,chen2020lottery,liang2021super,zafrir2021prune} and \emph{iterative pruning} \citep{han2015learning,zhu2017prune,paganini2020iterative,louizos2017learning,sanh2020movement}.
One-shot pruning specifies the sparsity pattern of a fully-trained dense model based on the weights' importance scores, and then trains a sparse model via ``rewinding'' (i.e., prune the fully-trained model and then re-train the pruned model). Such an approach has been widely utilized in exploring the Lottery Ticket Hypothesis \citep{frankle2018lottery,chen2020lottery,liang2021super}. However, determining the sparsity pattern based on fully-trained models fails to take the complicated training dynamics into consideration.
Some weights that are important to early stage of training do not necessarily have large importance scores in the fully-trained model, and therefore get pruned. Consequently, the rewinding process only yields models with worse generalization performance, especially in the high sparsity regime \citep{liang2021super}.

By contrast, iterative pruning conducts training and pruning simultaneously \citep{zhu2017prune}. Hence, the sparsity pattern can adapt to the parameter values and their dynamics throughout the entire training process. As such, iterative pruning attains significant improvement in the high sparsity regime over one-shot pruning \citep{sanh2020movement}.  
Popular iterative pruning methods consider two approaches: sparsity regularization \citep{louizos2017learning,behnke2021pruning} and cardinality constraint \citep{han2015learning,sanh2020movement}. The former applies sparsity-inducing regularization (e.g., $\ell_{0}$ or $\ell_{1}$ regularization) to shrink certain weights toward zero values. The later truncates undesired weights according to the rank of their importance scores.

Unfortunately, iterative pruning methods \citep{han2015learning,louizos2017learning,sanh2020movement} suffer from a significant drawback: a weight's importance score cannot accurately reflect its contribution to model performance. This is mainly due to two types of uncertainty during training: (i) As the amount of data is quite large, the importance scores at each iteration are computed based on a randomly selected small batch of the training data, and therefore are subject to large variance due to stochastic sampling; (ii) The weights' importance scores can drastically vary due to the complicated training dynamics and optimization choices (e.g., dropout). Therefore, some weights can frequently alternate between being pruned and being activated, which causes training instability or even divergence.

\begin{figure*}[t!]
	\centering
	\includegraphics[width=0.98\textwidth]{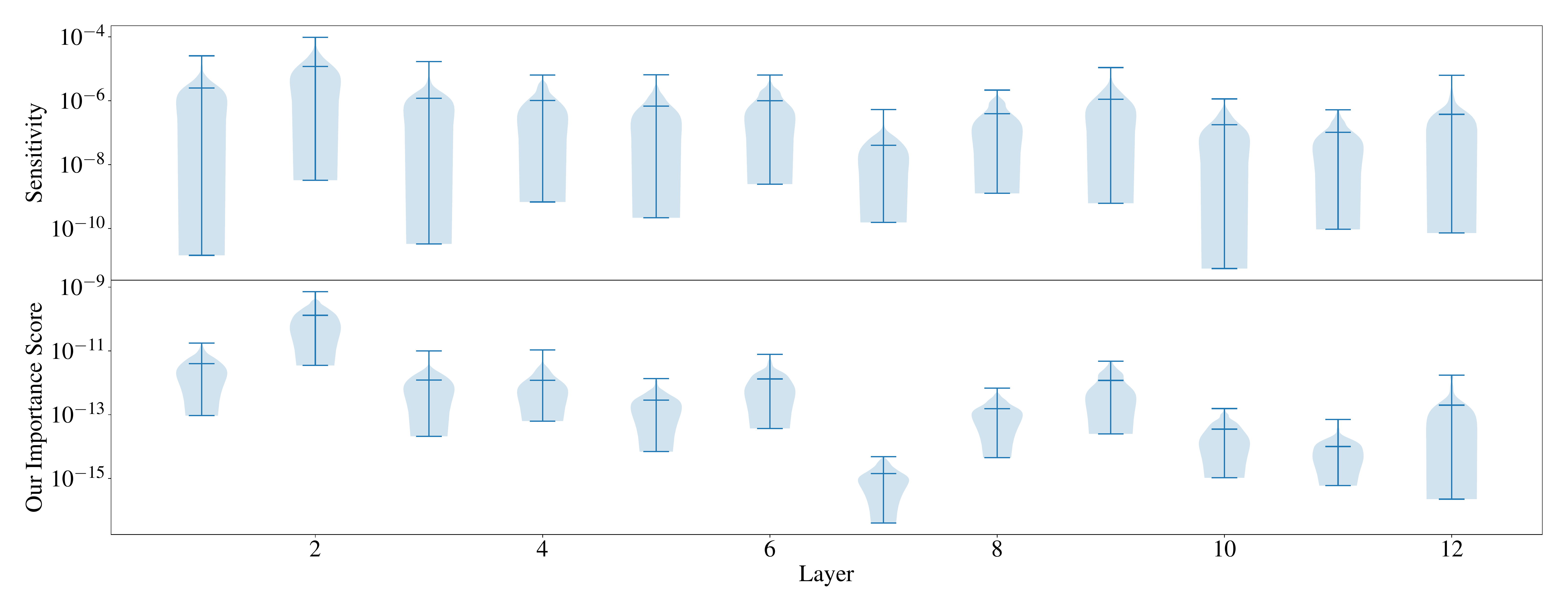}
	\vspace{-3mm}
	\caption{Variability of importance score when pruning BERT-base on RTE using iterative pruning. We compare two importance metrics: sensitivity in \eqref{eq:sensitivity} (top subfigure) and our proposed importance score in \eqref{eq:importance_score} (bottom subfigure). We calculate the importance scores every 10 steps and randomly sample 12 weights (one from each layer) for illustration purposes. Each violin plot corresponds to the importance scores' distribution of one sampled weight. Note that the plot is in \emph{log} scale. We remark the remaining weights behave similarly. 
	}
	\label{fig:sensitivity_distribution}
	 \vspace{-2mm}
\end{figure*}

As an example, Figure~\ref{fig:sensitivity_distribution} (top) illustrates variability of the weights' importance scores (referred to as sensitivity in the figure). We see that during training, importance scores of the sampled weights vary by order of magnitudes. This indicates that some weights indeed alternate between being important (retained) and unimportant (pruned).

To resolve this issue, we propose \emph{{\OurAlg}} (\underline{P}runing \underline{LA}rge \underline{T}ransf\underline{O}rmer with u\underline{N}certainty), which prunes the weights by considering both their importance and uncertainty (explained below). For the weights with low importance but high uncertainty, our method tends to retain them. More specifically, {\OurAlg} contains two important components: (i) We consider exponential moving average of the importance scores, which is a smoothed version of the metric. This encourages the model to retain weight whose importance scores abruptly drop due to training instability; (ii) We quantify uncertainty of importance estimation by its local temporal variation, which is defined as the absolute difference between the importance score at the current iteration and the score's exponential moving average of the previous iterations. 
Such quantification can be regarded as upper confidence bound (UCB) of estimated importance.  
A large local temporal variation implies a high uncertainty in the importance score at the current iteration. Therefore, it is not a reliable indicator of importance, and we should retain the weight even though its score is relatively small. 
In {\OurAlg}, we choose the product between the smoothed importance score and the local temporal variation as a new importance metric for the weights. 
From Figure~\ref{fig:sensitivity_distribution} (bottom), we see that variability of the proposed importance metric is drastically smaller.

We conduct extensive experiments on a wide range of tasks and models to demonstrate the effectiveness of {\OurAlg}.
We evaluate the performance of our method using BERT-base \citep{devlin2018bert} and DeBERTaV3-base \citep{he2021debertav3} on natural language understanding (GLUE, \citealt{wang2018glue}) and question answering (SQuAD, \citealt{rajpurkar2016squad}) datasets. 
We also apply our method to ViT-B16 \citep{dosovitskiy2020image} and evaluate the performance on CIFAR100 \citep{Krizhevsky09learningmultiple} and ImageNet \citep{deng2009imagenet}.
Experimental results show that {\OurAlg} significantly outperforms the baselines, especially under high sparsity settings (e.g., retaining only $10\%$ of the weights). For example, under $90\%$ sparsity on the MNLI dataset, our method achieves a $1.2\%$ accuracy improvement compared with state-of-the-art approaches.
 
\section{Background}\label{sec:preliminary}

We briefly review background and related works. 

\subsection{Pre-trained Transformer-based Models}\label{sec:pertrained_models}
Pre-trained Transformer-based models \citep{devlin2018bert,liu2019roberta,brown2020language,dosovitskiy2020image,he2021deberta,he2021debertav3} have manifested superior performance in various natural language processing and computer vision tasks. These models are often pre-trained on enormous amounts of unlabeled data in a unsupervised/self-supervised manner such that they can learn rich semantic knowledge. By further fine-tuning these pre-trained models, we can effectively transfer such knowledge to benefit downstream tasks.

\subsection{Importance Scores for Pruning}\label{sec:weight_importance} 

We denote $ \bTheta = [\theta_1, \theta_2, \dots, \theta_d] \in \RR^{d}$ parameters of a model, and further define $ \bThetaj = [0, \dots, 0, \theta_j , 0, \\ \dots, 0] \in \RR^{d}$. The importance score of $\bTheta$ is denoted as $ \Sc(\bTheta) \in \RR^d$, where its $j$-th index $\Sc_j(\bTheta)$ is the importance score for $\theta_j$. When the context is clear, we simply write $ \Sc $.

{\bf Parameter magnitude} is an effective importance metric for model pruning \citep{han2015learning,han2015deep,paganini2020iterative,zhu2017prune,renda2020comparing,zafrir2021prune}. It defines the importance of a weight as its magnitude, i.e., $ \Sc_{j} = |\theta_{j}|$. Parameters with small magnitude are pruned. Such a simple metric, however, cannot properly quantify a weight's contribution to the model output. This is because a small weight can yield a huge influence to the model output due to the complex compositional structure of neural networks, and vice versa.

{\bf Sensitivity of parameters} is another importance metric \citep{molchanov2019importance,sanh2020movement,liang2021super}. It essentially approximates the change in loss when a parameter is zeroed out. If the removal of a parameter causes a large influence on the loss, then the model is sensitive to it. 
More specifically, the sensitivity of the $j$-th parameter in $\bTheta$ is defined as the magnitude of the gradient-weight product: 
\begin{equation}\label{eq:sensitivity}
	\I_j = |\bThetaj^{\top} \nabla\cL(\bTheta)|.
\end{equation} 
This definition is derived from the first-order Taylor expansion of $ \cL(\cdot) $ with respect to $ \theta_j $: $ \I_j $ approximates the absolute change of the loss given the removal of $ \theta_j $,
\begin{align}\label{eq:sensitivity_taylor}
	\bThetaj^{\top} \nabla\cL(\bTheta) \approxeq \cL(\bTheta) - \cL(\bTheta - \bThetaj).
\end{align}

\subsection{Iterative Pruning} 

Iterative pruning methods \citep{han2015learning,louizos2017learning,sanh2020movement} have demonstrated to be effective in natural language processing and computer vision. 
These methods prune model weights based on the ranking of their importance scores, where weights with small scores are pruned.
Specifically, at the $t$-th iteration, we first take a stochastic gradient descent step,
\begin{align*}
\widetilde{\bTheta}^{(t)} = \bThetat - \alpha \nabla \cL(\bThetat),
\end{align*}
where $\alpha>0$ is the learning rate. Then, given the importance score $ \Sct $ (e.g., magnitude or sensitivity) for $\bTheta$, the weights are pruned following
\begin{align*}
\bTheta^{(t+1)} = \Proj(\widetilde{\bTheta}^{(t)}, \Sct),
\end{align*}
where the $j$-th entry of $\Proj(\widetilde{\bTheta}, \Sct)$ is defined as
\begin{align} \label{eq:proj_function}
[\Proj(\widetilde{\bTheta},\Sc)]_j = \left\{
\begin{array}{lc}
\widetilde{\bTheta}_j &\textrm{if}~\Sc_j~\textrm{is~in~the~top~$r\%$ of $\Sc$}, \\
0 &\textrm{otherwise}.
\end{array}
\right.
\end{align}
Here $\rr$ is a pre-defined parameter that determines the percentage of nonzero weights in the pruned model.

\section{Method}\label{sec:method}
We present our algorithm -- {\OurAlg} (\underline{P}runing \underline{LA}rge \underline{T}ransf\underline{O}rmer with u\underline{N}certainty) to overcome the drawbacks of existing pruning methods. Our proposed algorithm contains two important components: (i) sensitivity smoothing using exponential moving average, which reduces the non-negligible variability due to training dynamics and mini-batch sampling. (ii) uncertainty qualification using local temporal variation, which drives the algorithm to explore the potentially important weights.

\paragraph{Sensitivity Smoothing} 
In practice, the sensitivity in \eqref{eq:sensitivity} exhibits large variability because of two reasons.
First, $\I_j$ in \eqref{eq:sensitivity} is computed batch-wise, i.e., in each iteration we randomly sample a mini-batch and compute sensitivity of the weights using this batch.
Second, the complicated training dynamics and optimization settings (e.g., dropout) further amplify such variability.
To mitigate this issue, we propose to smooth $ \I_{j} $ by exponential moving average.
Concretely, at the $t$-th iteration, we have the smoothed sensitivity
\begin{equation}\label{eq:ema_sensitivity}
\Ibartj = \betaI \Ibarj^{(t-1)} + (1-\betaI)\Itj, 
\end{equation}
where $ \betaI \in (0, 1) $ is a hyper-parameter and  $\Ibarj^{(0)} = 0~\textrm{for~all}~j = 1 \dots d.$

The smoothed sensitivity $ \Ibarj $ can effectively reduce variability.
That is, for a weight $\theta_j$, an abrupt drop in its sensitivity $I_j$ causes limited impact on $\Ibarj$.
Therefore, training using the smoothed sensitivity is stable because the pruning algorithm outputs a stable sparsity pattern.
Furthermore, the exponential moving average emphasizes recent sensitivity scores and drops the stale information, which further benefits training.

\paragraph{Uncertainty Quantification} 
Besides sensitivity smoothing, we also directly consider the uncertainty of importance estimation to reduce variability.
Specifically, we quantify the estimation uncertainty by the sensitivity's local temporal variation, defined as
\begin{equation}\label{eq:uncertainty}
\Utj = |\Itj - \Ibartj|. 
\end{equation}
Similar to \eqref{eq:ema_sensitivity}, we further apply exponential moving averaging to $ \Utj $: 
\begin{equation}\label{eq:ema_uncertainty}
\Ubartj = \betaU \Ubarj^{(t-1)} + (1-\betaU) \Utj,
\end{equation}
where we set $\Ubarj^{(0)}=0$ for all $j=1\cdots d$.
The uncertainty quantifies the variability by considering the difference between the current sensitivity and its historical average.
A large $ \Ubartj $ indicates that the sensitivity $I_j^{(t)}$ computed using the current batch significantly deviates from the weight's historical sensitivity $\Ibarj^{(t)}$.
This implies that there exists high uncertainty in $ \Ibartj $, and hence, $\Ibarj^{(t)}$ is not yet a reliable indicator of the importance of $\theta_j$. 
In this sense, $\Ubartj$ can be regarded as an upper confidence bound of estimated importance $\Ibartj$. 

\begin{algorithm}[htb!]
	\caption{{\OurAlg}} 
	\label{alg:our_algorithm}
	\begin{algorithmic}[1]
		\STATE {\bfseries Input:} Dataset $\DD$; total training iterations $T$; exponential moving average parameters $\betaI$ and $\betaU$; remaining weights ratio $\rr$; learning rate $\alpha$. 
		\STATE \textbf{Initialize:} $\Ubar^{(0)}=0$, $\Ibar^{(0)}=0$;
		\FOR{$ t = 0, \dots, T-1 $}
		\STATE Sample a mini-batch from $\mathcal{D}$;
		\STATE Compute the gradient $ \nabla \cL(\bThetat) $;
		\STATE Compute $ \Itj = |{\bTheta}^{(t) \top}_{j,-j} \nabla \cL(\bThetat)| $; 
		\STATE Compute $ \Ibartj = \betaI \Ibarj^{(t-1)} + (1-\betaI)\Itj $;
		\STATE Compute $ \Utj = |\Itj - \Ibartj| $;
		\STATE Compute $ \Ubartj = \betaU \Ubarj^{(t-1)} + (1-\betaU) \Utj  $;
		\STATE Compute $ \Sc^{(t)} = \Ibart \odot \Ubart$;
		\STATE Update $ \bTheta^{(t+1)} = \Proj ( \bThetat - \alpha \nabla \cL(\bThetat) , \Sc^{(t)})$;
		\ENDFOR
		\STATE \textbf{Output:} Pruned model $\bTheta^{(T)}$.
	\end{algorithmic}
\end{algorithm}

\paragraph{Algorithm} 

The most important difference between our proposed algorithm and existing ones is the importance score, which is defined as: 
\begin{equation}\label{eq:importance_score}
\Sct = \Ibart \odot \Ubart,
\end{equation}
where $ \odot $ is Hadamard product.
As can be seen from \eqref{eq:importance_score}, $\Ibart$ measures the sensitivity of weights while $ \Ubart $ quantities the uncertainty of sensitivity estimation. 
The product yields a trade-off between $\Ibart$ and $\Ubart$. Specifically, when a weight $\theta_j$ has a high uncertainty $\Ubart$, even though its sensitivity $\Ibart$ at the current iteration is low, it may still significantly increase due to the high variability introduced by the mini-batch sampling and complicated training dynamics. 
Therefore, we make a conservative choice and our proposed algorithm tends to retain it and explore it for a longer time.

\begin{wrapfigure}{r}{65mm}
    \centering
    \vspace{-8mm}
   	\includegraphics[width=0.42\textwidth]{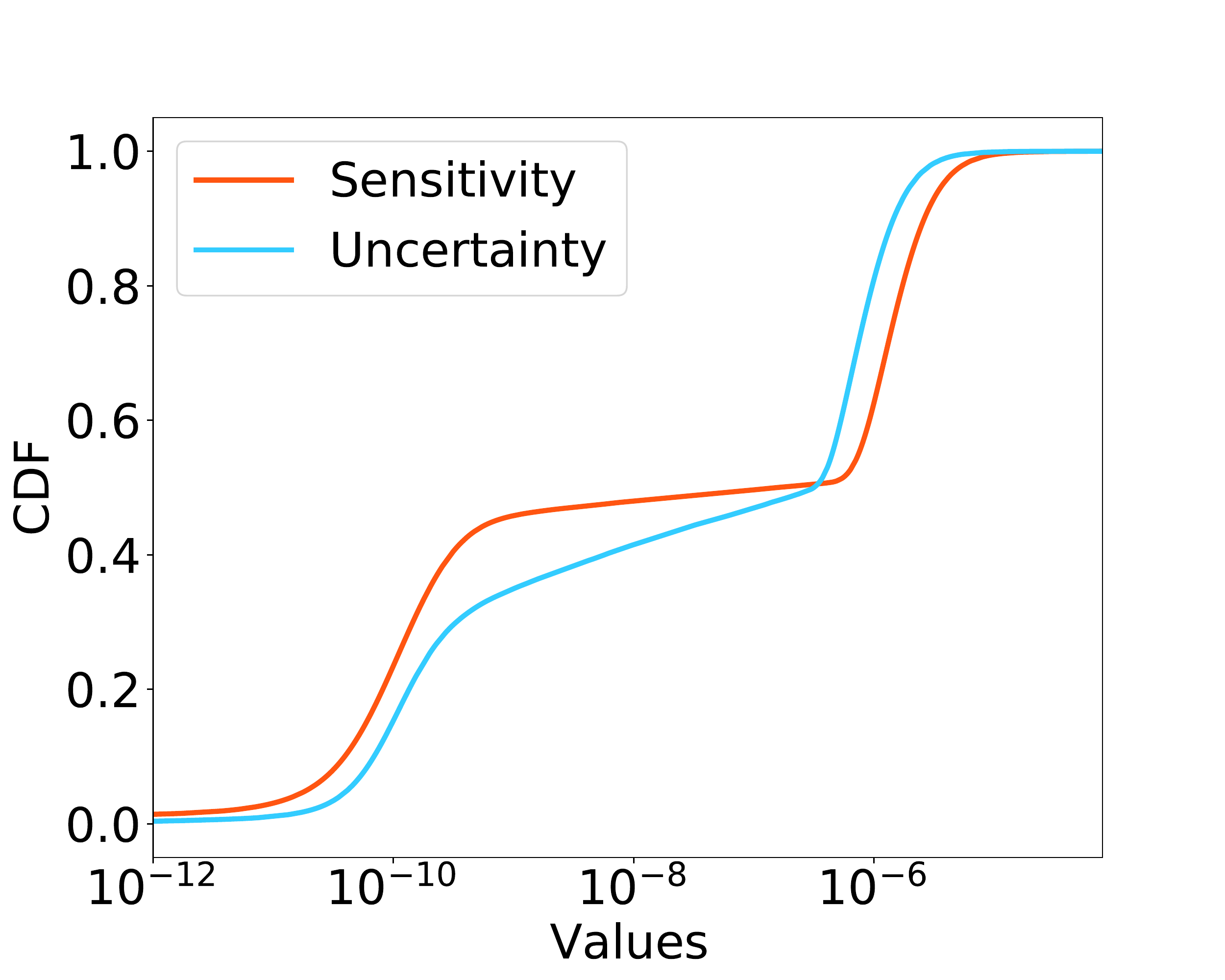}
   	\vspace{-10mm}
    \caption{\small The cumulative distribution function (CDF) of $\Ibartj$ and $\Ubartj$ over $j$ at $t=500$ when pruning BERT-base on RTE. }
    \label{fig:cdf}
\end{wrapfigure}

We remark \eqref{eq:importance_score} shares the same spirit with UCB methods for bandit problems \citep{lai1985asymptotically,zhang2021biased}. Each parameter is considered as an arm, $\Ibartj$ is regarded as estimated rewards and $\Ubartj$ controls the level of uncertainty. 
Since $ \Ibartj $ and $ \Ubartj $ are highly skewed to zero as shown by Figure~\ref{fig:cdf}, \eqref{eq:importance_score} applies a logarithmic transformation to make them distribute more evenly: 
\begin{align*}
    \Sct_{j} = \exp\Big( \log(\Ibartj) + c \log(\Ubartj) \Big),
\end{align*}
where we let $c=1$. The above equation aligns to the policy of UCB. The upper confidence bound $\Ubartj$ quantifies the uncertainty of importance estimation and promotes the exploration.

With our defined importance score, our algorithm prunes the model weights after each gradient decent step, i.e.,
\begin{equation}\label{eq:pruning_alg}
\bTheta^{(t+1)} = \Proj ( \bThetat - \alpha \nabla \cL(\bThetat) , \Sc^{(t)}),
\end{equation}
where $\Proj$ is defined in \eqref{eq:proj_function}.
We summarize our proposed algorithm \emph{{\OurAlg}} in Algorithm~\ref{alg:our_algorithm}.

\section{Experiments}\label{sec:experiments}
We implement {\OurAlg} for pruning pre-trained BERT-base \cite{devlin2018bert}, DeBERTaV3-base \cite{he2021deberta} and ViT-B16 \cite{dosovitskiy2020image} during fine-tuning.
We evaluate effectiveness of the proposed algorithm on natural language understanding (GLUE, \citealt{wang2018glue}), question answer (SQuAD v1.1, \citealt{rajpurkar2016squad}) and image classification (CIFAR100, \citealt{krizhevsky2009learning}; ImageNet, \citealt{deng2009imagenet}) tasks.
All the gains have passed significance tests with $p<0.05$.

\textbf{Implementation details.}
We use \textit{PyTorch} \citep{paszke2019pytorch} to implement all the algorithms.
Our implementation is based on the publicly available \textit{MT-DNN} \citep{liu2019multi,mtdnn2020demo}\footnote{\url{https://github.com/microsoft/MT-DNN}} and \textit{Huggingface Transformers}\footnote{\url{https://github.com/huggingface/transformers}} \citep{wolf2019huggingface} code-base.
All the experiments are conducted on NVIDIA V100 GPUs.

In Algorithm~\ref{alg:our_algorithm}, a weight may be zeroed-out (pruned) and reactivated in later iterations.
In this case, we let the weight starts from zero instead of from the value before zeroing-out.
This is because a pruned weight is still included in the training dynamics as a zero value, such that its is natural to restart it from zero.

We use a cubic schedule \cite{zhu2017prune,sanh2020movement,zafrir2021prune} to gradually increase the sparsity level during pruning. Please refer to Appendix~\ref{app:schedule} for details.

\vspace{0.05in} \noindent
{\bf Baselines.}
We compare {\OurAlg} with the following methods:

$\bullet$
\textit{$\ell_0$ regularization} \citep{louizos2017learning} is an effective modeling pruning method. The method adds a penalty to the proportion of remaining weights.

$\bullet$
\textit{Magnitude pruning} \citep{zhu2017prune} propose an automated gradual pruning method, which is a magnitude-based pruning method. The method enables masked weights to be updated, and has achieved state-of-the-art results among magnitude-based approaches.

$\bullet$
\textit{Movement pruning} \citep{sanh2020movement} is a state-of-the-art method for model pruning. The method applies an iterative pruning strategy, where sensitivity is used as the importance metric. The approach considers the changes in weights during training, and has achieved superior performance.

$\bullet$
\textit{Soft movement pruning} \citep{sanh2020movement} is a soft version of movement pruning based on the binary mask function in \citet{mallya2018piggyback}.

We use publicly available implementation\footnote{\url{https://github.com/huggingface/transformers/tree/master/examples/research_projects/movement-pruning}} to run all the baselines. Please refer to \citet{sanh2020movement} and references therein for more details.

\setlength{\tabcolsep}{0.38em}
\begin{table*}[t!]
	\caption{Results with BERT-base on GLUE development set. Here \textit{Ratio} is the proportion of remaining weights. Results with \emph{N.A.} indicate the model does not converge. The best results on each dataset are shown in \textbf{bold}. We report mean of $5$ runs using different random seeds.}
	\vspace{-3mm}
	\label{tab:glue_datasets}
	\begin{center}
		\begin{small}
			\begin{tabular}{l|l|cccccccc}
				\toprule
				\multirow{2}*{\bf Ratio} & \multirow{2}*{\bf Method} & {\bf MNLI} & {\bf RTE} & {\bf QNLI}  & {\bf MRPC} & {\bf QQP } & {\bf SST-2} & {\bf CoLA} & {\bf STS-B} \\  
				~ & ~ & {m / mm} & {Acc} & {Acc} & {Acc / F1} & {Acc / F1} & {Acc} & {Mcc} & { P/S Corr} \\
				\midrule 
				{\bf 100\%} & {$\text{BERT}_{\text{base}}$} & {84.6 / 83.4} & {69.3} & {91.3} & {86.4 /  90.3} & {91.5 / 88.5} & {92.7} & {58.3} & {90.2 / 89.7}
				\\
				\midrule
				\multirow{5}*{\bf 20\%} & {$ \ell_{0} $ Regularization} & {80.5 / 81.1} & {63.2} & {85.0} & {75.7 / 80.2} & {88.5 / 83.3} & {85.0} & \emph{N.A.} & {82.8 / 84.7} \\
				~ & {Magnitude} & {81.5 / 82.9} & {65.7} & {89.2} & {79.9 / 86.2}  & {86.0 / 83.8} & {84.3} & {42.5} & {86.8 / 86.6} \\
				~ & {Movement} & {80.6 / 80.8} & \emph{N.A.} & {81.7} & {68.4 / 81.1} & {89.2 / 85.7} & {82.3} & \emph{N.A.} & \emph{N.A.} 
				\\
				~ & {Soft-Movement} & {81.6 / 82.1} & {62.8} & {88.3} & {80.9 / 86.7} & {90.6 / \textbf{87.5}} & {89.0} & {48.5} & {87.8 / 87.5}  \\
				~ & {\OurAlg} & {\bf83.1 / 83.4} & {\bf68.6} & {\bf90.1} & {\bf85.5 / 89.8} & {\bf90.7 / 87.5} & {\bf91.3} & {\bf54.5} & {\bf89.0 / 88.5}  
				\\
				\midrule
				\multirow{5}*{\bf 15\%} & {$ \ell_{0} $ Regularization} & {79.1 / 79.8} & {62.5} & {84.0} & {74.8 / 79.8} & {87.9 / 82.3} & {82.8} & \emph{N.A.} & {81.8 / 84.2} \\
				~ & {Magnitude} & {80.1 / 80.7} & {64.6} & {88.0} & {69.6 / 79.4} & {83.6 / 79.2} & {82.8} & \emph{N.A.} & {85.4 / 85.0}
				\\
				~ & {Movement} & {80.1 / 80.3} & \emph{N.A.} & {81.2} & {68.4 / 81.0} & {89.6 / 86.1} & {81.8} & \emph{N.A.} & \emph{N.A.}
				\\
				~ & {Soft-Movement} & {81.2 / 81.7} & {60.2} & {87.2} & {81.1 / 87.0} & {90.4 / 87.1} & {88.4} & {40.8} & {86.9 / 86.6}  \\ 
				~ & {\OurAlg} & {\bf82.7 / 83.0} & {\bf65.7} & {\bf89.9} & {\bf85.3 / 89.5} & {\bf90.5 / 87.3} & {\bf91.1} & {\bf52.5} & {\bf88.4 / 87.9} 
				\\
				\midrule
				\multirow{5}*{\bf 10\%} & {$ \ell_{0} $ Regularization} & {78.0 / 78.7} & {59.9} & {82.8} & {73.8 / 79.5} & {87.6 / 82.0} & {82.5} & \emph{N.A.} & {82.7 / 83.9} \\
				~ & {Magnitude} & {78.8 / 79.0} & {57.4} & {86.6} & {70.3 / 80.3}  & {78.8 / 77.0} & {80.7} & \emph{N.A.} & {83.4 / 83.3} \\
				~ & {Movement} & {79.3 / 79.5} & \emph{N.A.} & {79.2} & {68.4 / 81.2} & {89.1 / 85.4} & {80.2} & \emph{N.A.} & \emph{N.A.} \\
				~ & {Soft-Movement} & {80.7 / 81.1} & {58.8} & {86.6} & {79.7 / 85.9} & {\textbf{90.2} / 86.7} & {87.4} & \emph{N.A.} & {86.5 / 86.3}  \\
				~ & {\OurAlg} & {\bf82.0 / 82.2} & {\bf65.3} & {\bf88.9} & {\bf84.3 / 88.8} & {\bf90.2 / 86.8} & {\bf90.5} & {\bf44.3} & {\bf87.4 / 87.1} \\
				\bottomrule
			\end{tabular}
		\end{small}
	\end{center}
\end{table*}

\subsection{Natural Language Understanding}\label{sec:glue_expriments}
{\bf Models and Datasets.} We evaluate the pruning performance of BERT-base \citep{devlin2018bert} and DeBERTaV3-base \citep{he2021deberta} using the proposed algorithm.
We conduct experiments on the General Language Understanding Evaluation (GLUE, \citealt{wang2018glue}) benchmark.
The benchmark includes two single-sentence classification tasks: SST-2 \citep{sst2013} and CoLA \citep{cola2018}. 
GLUE also contains three similarity and paraphrase tasks: MRPC \citep{mrpc2005}, STS-B \citep{sts-b2017} and QQP.
There are also four natural language inference tasks in GLUE: MNLI \citep{mnli2018}; QNLI \citep{squad1}; RTE \citep{rte1, rte2, rte3, rte5}; and WNLI \citep{wnli2012}.
Following previous works, we exclude WNLI in the experiments.
Dataset details are summarized in Appendix~\ref{app:dataset-glue}.

\vspace{0.05in} \noindent
{\bf Implementation Details.}
We select the exponential moving average parameters $ \betaI $ from $\{0.75, 0.80, \\ 0.85, 0.90\}$ and $ \betaU $ from $ \{0.850, 0.900, 0.950, 0.975\} $. We select the learning rate from $\{3\times 10^{-5}, 5\times 10^{-5}, 8\times 10^{-5}, 1\times 10^{-4}  \}$ and the batch size from $\{ 8, 16, 32 \}$. More details are presented in Appendix~\ref{sec:app_nlu}.   

\vspace{0.05in} \noindent
{\bf Main results. }
We compare {\OurAlg} with the baseline methods under different sparsity levels: $90\%$, $85\%$ and $80\%$.
Table~\ref{tab:glue_datasets} shows experimental results on the GLUE development set. We see that {\OurAlg} achieves better or on par performance compared with existing approaches on all the datasets under all the sparsity levels.
For example, when the sparsity level is $80\%$, {\OurAlg} achieves a $83.1\%$ accuracy on the MNLI-m dataset, which is $1.5\%$ higher than the best-performing baseline (soft movement pruning).
The performance gain of our method is even more significant on small datasets. For example, {\OurAlg} outperforms existing approaches by more than $5\%$ on RTE when the sparsity level is $80\%$.
We remark that while movement pruning and its soft version perform well on large datasets (e.g., MNLI and QQP), it behaves poorly or even cannot converge on small datasets (e.g., RTE, CoLA and STS-B). We provide plausible explanations in Section~\ref{sec:discussion}.

Table~\ref{tab:deberta_experiemnts} summarizes experimental results on MNLI and SST-2 for pruning DeBERTaV3-base. Similar to Table~\ref{tab:glue_datasets}, {\OurAlg} significantly outperforms the baseline method on all the datasets under all the sparsity levels. From the results, we see that for the DeBERTa model, our method obtains more performance gain when the sparsity level is high. For example, in the $70\%$ sparse case, {\OurAlg} outperforms the baseline by $1.9\%$ ($94.6$ v.s. $92.7$) on the SST-2 dataset; while in the $90\%$ sparse case, our method achieves a $6.6\%$ improvement ($90.0$ v.s. $83.4$).

\setlength{\tabcolsep}{0.38em}
\begin{table}[htb!]
	\caption{Results with DeBERTaV3-base on SQuAD v1.1, MNLI and SST-2. Here \textit{Ratio} is the proportion of remaining weights. The best results on each dataset are shown in \textbf{bold}.}
	\vspace{-3mm}
	\label{tab:deberta_experiemnts}
	\begin{center}
		\begin{small}
			\begin{tabular}{c|c|ccc}
				\toprule
				\multirow{2}*{\bf Ratio} & \multirow{2}*{\bf Method} & {\bf MNLI} & {\bf SST-2} & {\bf SQuAD} \\  
				~ & ~ & {m / mm} & {Acc}  & { EM / F1} \\
				\midrule 
				{\bf 100\%} & {$\text{BERT}_{\text{base}}$} & {89.9 / 90.1}  & {95.6} & {84.6 / 92.0} \\
				\midrule
				\multirow{2}*{\bf 30\%} & {Magnitude} & {87.6 / 87.3} & {92.7} & {82.2 / 89.9} \\ 
				~ & {{\OurAlg}} & {\bf 88.9 / 88.8} & {\bf 94.6} & {\bf 83.1 / 90.9} \\
				\midrule
				\multirow{2}*{\bf 20\%} & {Magnitude} & {82.3 / 83.8} & {90.8} & {78.8 / 86.7} \\ 
				~ & {{\OurAlg}} & {\bf 87.2 / 87.0} & {\bf93.1} & {\bf81.9 / 89.8} \\
				\midrule
				\multirow{2}*{\bf 15\%} & {Magnitude} & {80.7 / 81.0} & {88.5} & {75.8 / 84.6} \\ 
				~ & {{\OurAlg}} & {\bf 85.8 / 85.9} & {\bf 92.3} & {\bf 81.2 / 89.0}  \\
				\midrule
				\multirow{2}*{\bf 10\%} & {Magnitude} & {77.1 / 78.0} & {83.4} & {70.5 / 80.5} \\ 
				~ & {{\OurAlg}} & {\bf 83.4 / 83.5} & {\bf 90.0} & {\bf 79.0 / 87.1} \\
				\bottomrule
			\end{tabular}
		\end{small}
	\end{center}
	\vspace{-3mm}
\end{table}

\subsection{Question Answering} 

{\bf Models and Datasets.}
We evaluate performance of the proposed algorithm on a question answering dataset (SQuAD v1.1, \citealt{squad1}), where we use {\OurAlg} to prune BERT-base and DeBERTaV3-base.
Question answering is treated as a sequence labeling problem, where we predict the probability of each token being the start and end of the answer span. The dataset contains $88k$ training and $10k$ validation samples.

\vspace{0.05in} \noindent
{\bf Implementation Details.}  We set the batch size as $16$, and the number of epochs for fine-tuning as $10$.
We use AdamW \citep{loshchilov2017decoupled} as the optimizer and we set the learning rate as $ 3\times 10^{-5} $. Please refer to Appendix~\ref{sec:app_squad} for more details.  

\vspace{0.05in} \noindent
{\bf Main Results. } 
Table~\ref{tab:squad_experiemnts} summarizes experimental results, where we prune a pre-trained BERT-base model under $6$ different sparsity settings. From the results, we see that {\OurAlg} consistently outperforms existing approaches under all sparsity levels in terms of the two evaluation metrics: exact match (EM) and F1.
Notice that movement pruning and soft movement pruning are more efficient in the high-sparsity regime (e.g., $90\%$ sparse); while in the low-sparsity regime (e.g., $50\%$ sparse), these methods behave on par or worse than magnitude pruning and $\ell_0$ regularization. Our method, on the other hand, is effective under all sparsity levels.
Also note that {\OurAlg} is more effective in the high-sparsity regime. For example, in the $50\%$ sparse case, our method outperforms the best-performing baseline (magnitude pruning) by $0.2\%$ in terms of F1; while in the $85\%$ sparse case, {\OurAlg} achieves a $2.5\%$ gain.

Table~\ref{tab:deberta_experiemnts} summarizes results of pruning DeBERTaV3-base on the SQuAD v1.1 dataset. Similar to the findings in Table~\ref{tab:squad_experiemnts}, {\OurAlg} significantly outperforms the baseline method. Additionally, our method is also more effective in the high-sparsity regime when pruning DeBERTa models.

\begin{table*}[htb!]
	\caption{Results with $\text{BERT}_{\text{base}}$ on SQuAD v1.1. We report EM/F1. Here \textit{Ratio} is the proportion of remaining weights. The best results in each setting are shown in \textbf{bold}.}
	\vspace{-3mm}
	\label{tab:squad_experiemnts}
	\begin{center}
		\begin{tabular}{l|cccccc}
			\toprule
			{\bf Ratio} & {\bf 10\%} & {\bf 15\%} & {\bf 20\%} & {\bf 30\%} & {\bf 40\%} & {\bf 50\%} \\
			\midrule 
			$\text{BERT}_{\text{base}}$ & \multicolumn{6}{|c}{80.4 / 88.1} \\ \midrule
			{$ \ell_{0} $ Regularization} & {68.9 / 80.0} & {70.7 / 80.9} & {72.0 / 81.9} & {73.1 / 82.8} & {74.3 / 83.9} & {75.1 / 84.6}  \\
			{Magnitude} & 67.7 / 78.2 & 73.4 / 82.9 & 75.9 / 84.8 & 77.4 / 86.5 & 77.9 / 86.7 & \textbf{78.5} / 87.0 \\
			{Movement}  & 71.9 / 81.9 & 72.1 / 81.8 & 72.3 / 82.0 & 71.6 / 81.9 & 72.7 / 82.8 & 73.4 / 83.0 \\
			{Soft-Movement}   & 71.0 / 81.0 & 72.2 / 82.2 & --    & 75.3 / 84.6 & --   & 77.0 / 85.8 \\
			\midrule
			{\OurAlg} & {\bf 74.2 / 83.8} & {\bf 75.9 / 85.4} & {\bf 76.8 / 86.1} & {\bf 77.5 / 86.7} & {\bf 78.0 / 86.9 } & {\bf 78.5 / 87.2}
			\\
			\bottomrule
		\end{tabular}
	\end{center}
	\vspace{-5mm}
\end{table*}

\subsection{Image Classification} 
{\bf Models and Datasets.}
We apply {\OurAlg} to prune a ViT-B16 model \citep{dosovitskiy2020image}, and we evaluate model performance on two image classification datasets: CIFAR100 \cite{krizhevsky2009learning} and ImageNet \cite{deng2009imagenet}.

\vspace{0.05in} \noindent
{\bf Implementation Details.} We implement ViT using the following codebase\footnote{\url{https://github.com/jeonsworld/ViT-pytorch}}.
We use SGD with momentum \cite{qian1999momentum} as the optimizer. For CIFAR100, we set the batch size as $512$ and the learning rate as $0.03$. For ImageNet, we set the batch size as $150$ and the learning rate as $0.03$. Detailed settings are deferred to Appendix~\ref{sec:app_vit}.

\vspace{0.05in} \noindent
{\bf Main Results.}  
Experimental results are summarized in Table~\ref{tab:vit_experiemnts}. We see that our method significantly outperforms existing methods on both the datasets (CIFAR100 and ImageNet) under all the sparsity levels. For example, the accuracy of {\OurAlg} is $76.3\%$ on ImageNet when the sparsity level is $80\%$, whereas the accuracy of magnitude pruning and movement pruning is only $66.5\%$ and $66.6\%$, respectively. Note that performance gain of our method is more significant in the high sparsity regime.

	\begin{table}[htb!]
		\caption{Results with ViT-B16 on CIFAR100 and ImageNet. Here \textit{Ratio} is the proportion of remaining weights. The best results on each dataset are shown in \textbf{bold}.}
		\vspace{-2mm}
		\label{tab:vit_experiemnts}
		\begin{center}
			\begin{tabular}{c|c|@{\hskip4pt}c@{\hskip4pt}@{\hskip4pt}c@{\hskip4pt}@{\hskip4pt}c@{\hskip4pt}|@{\hskip4pt}c@{\hskip4pt}}
				\toprule
				~ & {\bf Ratio} & {Magnitude} & {Movement} & {{\OurAlg}} & {\small ViT-B16} \\
				\midrule 
				\multirow{3}*{\rotatebox{90}{\small CIFAR100}} & {30\%} & {88.9} & {78.5} & {\bf90.1} & \multirow{3}*{92.3} \\
				~ & {20\%} & {84.6} & {78.6} & {\bf87.3} & ~ \\
				~ & {10\%} & {64.3} & {77.5} & {\bf81.2} \\
				\midrule
				\multirow{3}*{\rotatebox{90}{\small ImageNet}} & {40\%} & {81.5} & {69.3} & {\bf82.6} & \multirow{3}*{83.5} \\
				~ & {30\%} & {78.9} & {68.6} & {\bf80.5} \\
				~ & {20\%} & {66.5} & {66.6} & {\bf76.3} \\
				\bottomrule
			\end{tabular}
		\end{center}
		\vspace{-5mm}
	\end{table}

\subsection{Analysis}

\vspace{-2mm}
\begin{figure*}[htb!]
	\centering
	\begin{subfigure}{0.32\textwidth}
		\centering
		\includegraphics[width=1.0\textwidth]{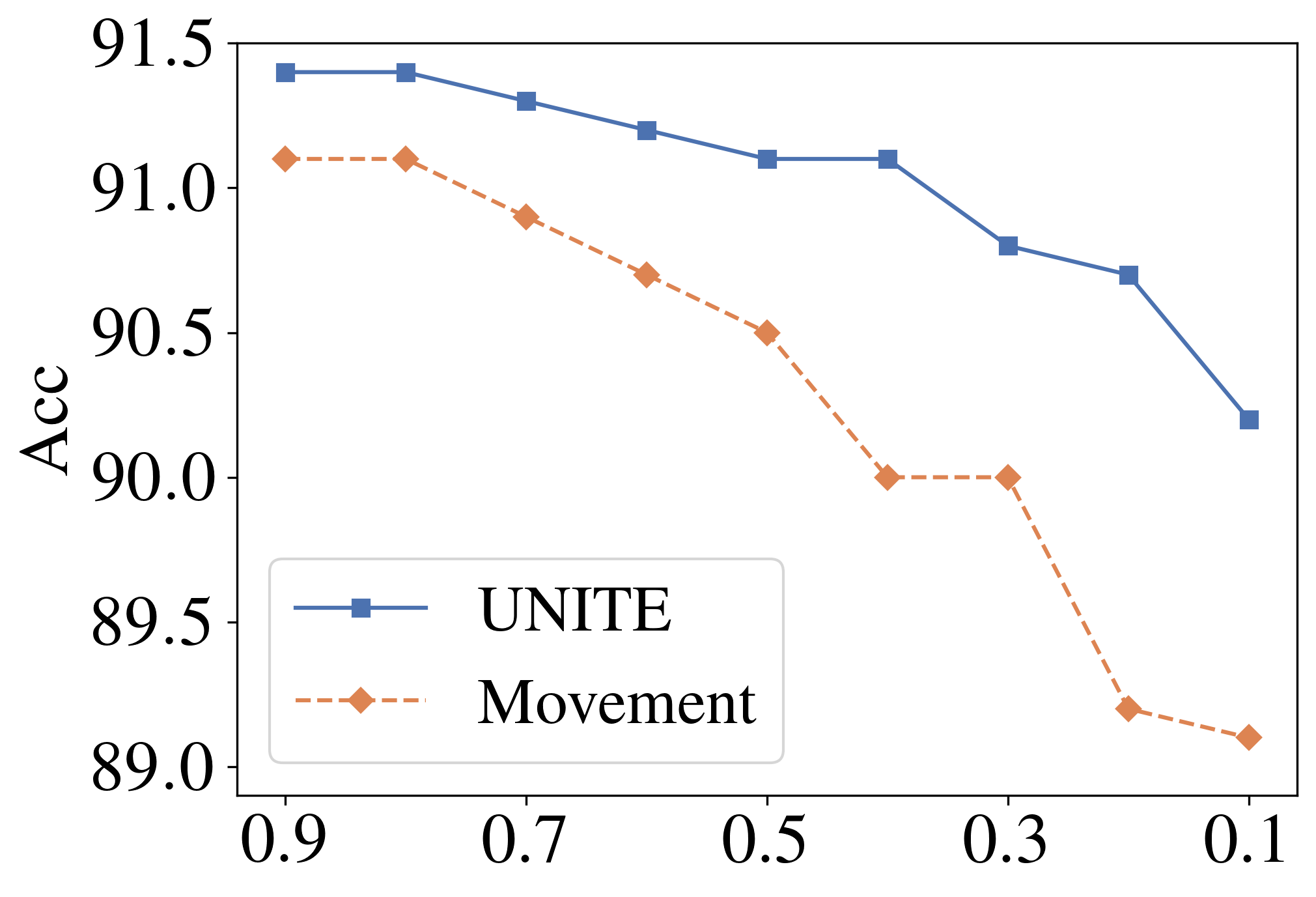}
		\vspace{-5mm}
		\caption{QQP}
	\end{subfigure}
	\begin{subfigure}{0.32\textwidth}
		\centering
		\includegraphics[width=1.0\textwidth]{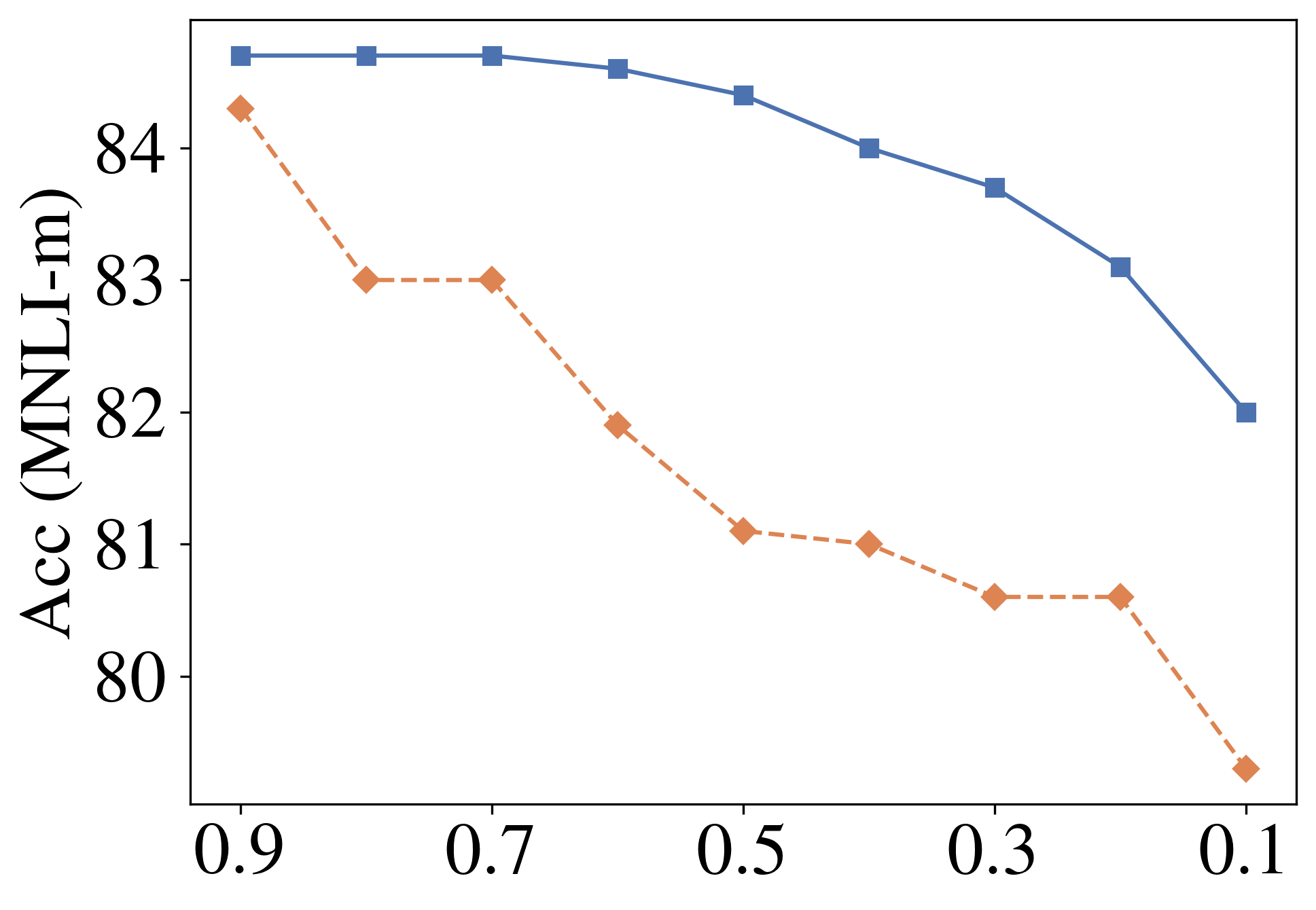}
		\vspace{-5mm}
		\caption{MNLI-m}
	\end{subfigure}
	\begin{subfigure}{0.32\textwidth}
		\centering
		\includegraphics[width=1.0\textwidth]{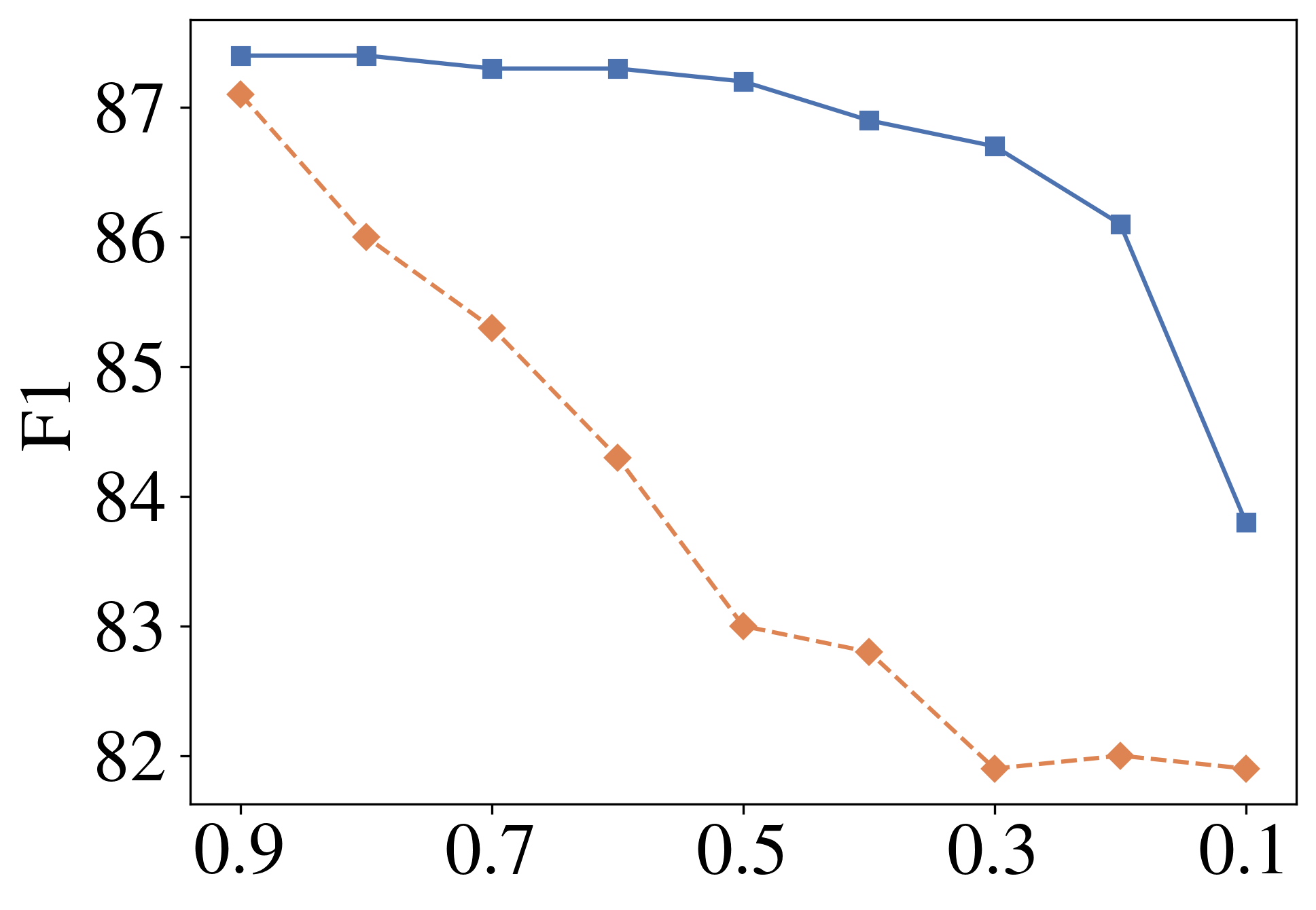}
		\vspace{-5mm}
		\caption{SQuAD v1.1}
	\end{subfigure}
	\vspace{-3mm}
	\caption{Model performance under different pruning ratio. We prune a BERT-base model during fine-tuning. Here the $x$-axis is the proportion of remaining weights.}
	\label{fig:pruning-ratio}
	\vspace{-2mm}
\end{figure*}

\textbf{Different levels of pruning ratio.}
Figure~\ref{fig:pruning-ratio} illustrates experimental results of pruning a BERT-base model during fine-tuning under different sparsity levels. We see that on all the three datasets (QQP, MNLI-m and SQuAD v1.1), {\OurAlg} achieves consistent performance improvement under all the sparsity levels compared with the baseline. Note that the performance gain is more significant when the sparsity level is high ($>50\%$). For example, {\OurAlg} outperforms the baseline by $0.4\%$ on MNLI-m when the sparsity level is $10\%$; while our method achieves a more than $3.0\%$ gain in accuracy when the sparsity level is $70\%$.

\vspace{0.1in} \noindent
\textbf{Variants of the importance score.}
Recall that in {\OurAlg}, the importance score is the product of sensitivity and uncertainty ($\Sc=\Ibar \odot \Ubar$). In Table~\ref{tab:importance}, we examine variants of the importance score. From the results, we see that using only the sensitivity ($\Sc=\Ibar$) or the uncertainty ($\Sc=\Ubar$) deteriorates model performance by over $1.0\%$ on all the three datasets.
We also examine the case where $\Sc=\Ibar/\Ubar$, i.e., we prune the weights with large uncertainty. In this case, the model fails to converge on CoLA, and the model performance drastically drops on the other datasets (i.e., $8.6\%$ on RTE and $9.6\%$ on SST-2). This indicates that weights with high uncertainty should be retained and further explored.

\begin{table*}[thb!]
	\centering
	\hspace{-4.5mm}
	\begin{minipage}{0.6\textwidth}
		\includegraphics[width=0.45\textwidth]{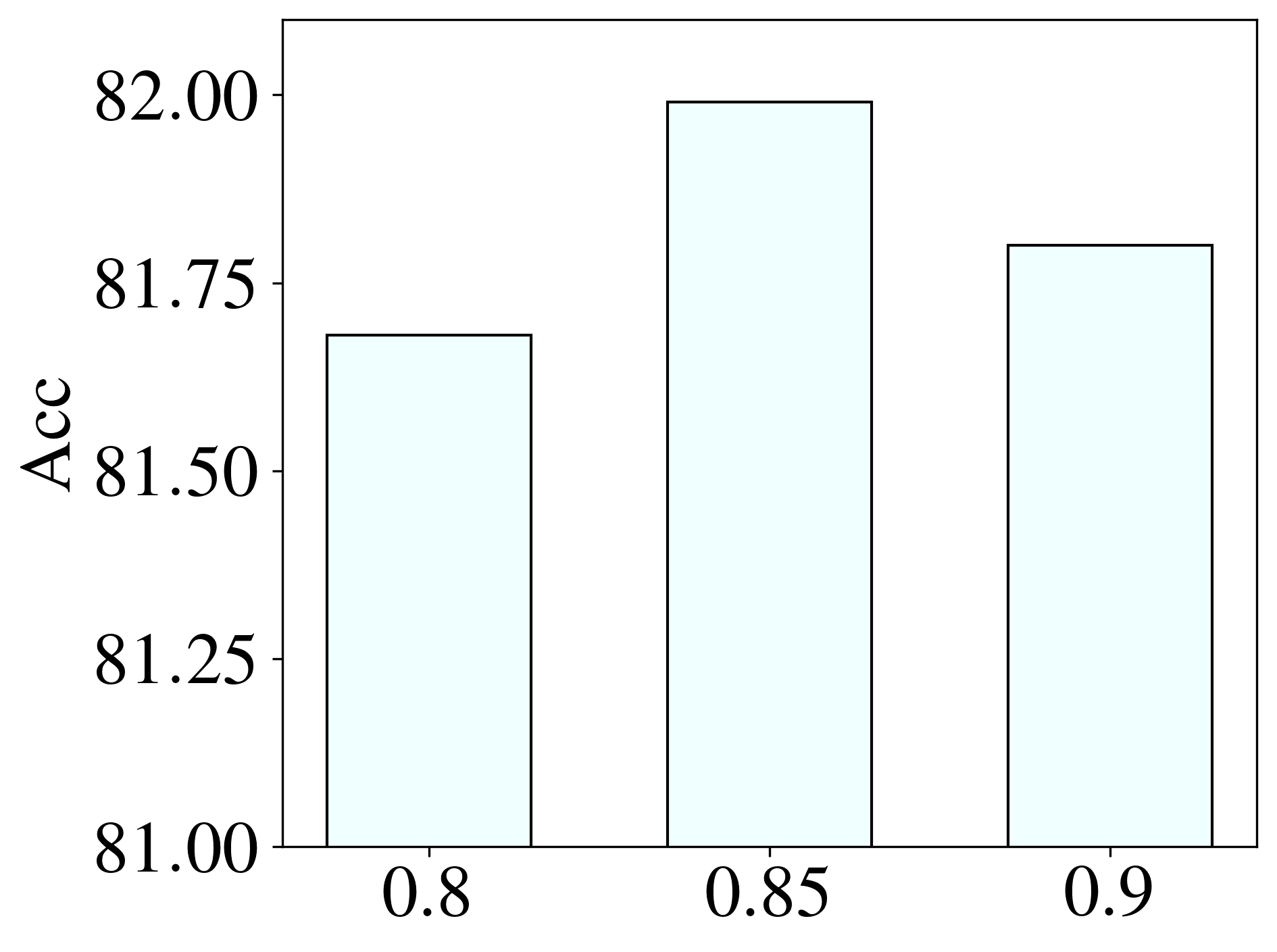} \hfill
		\includegraphics[width=0.45\textwidth]{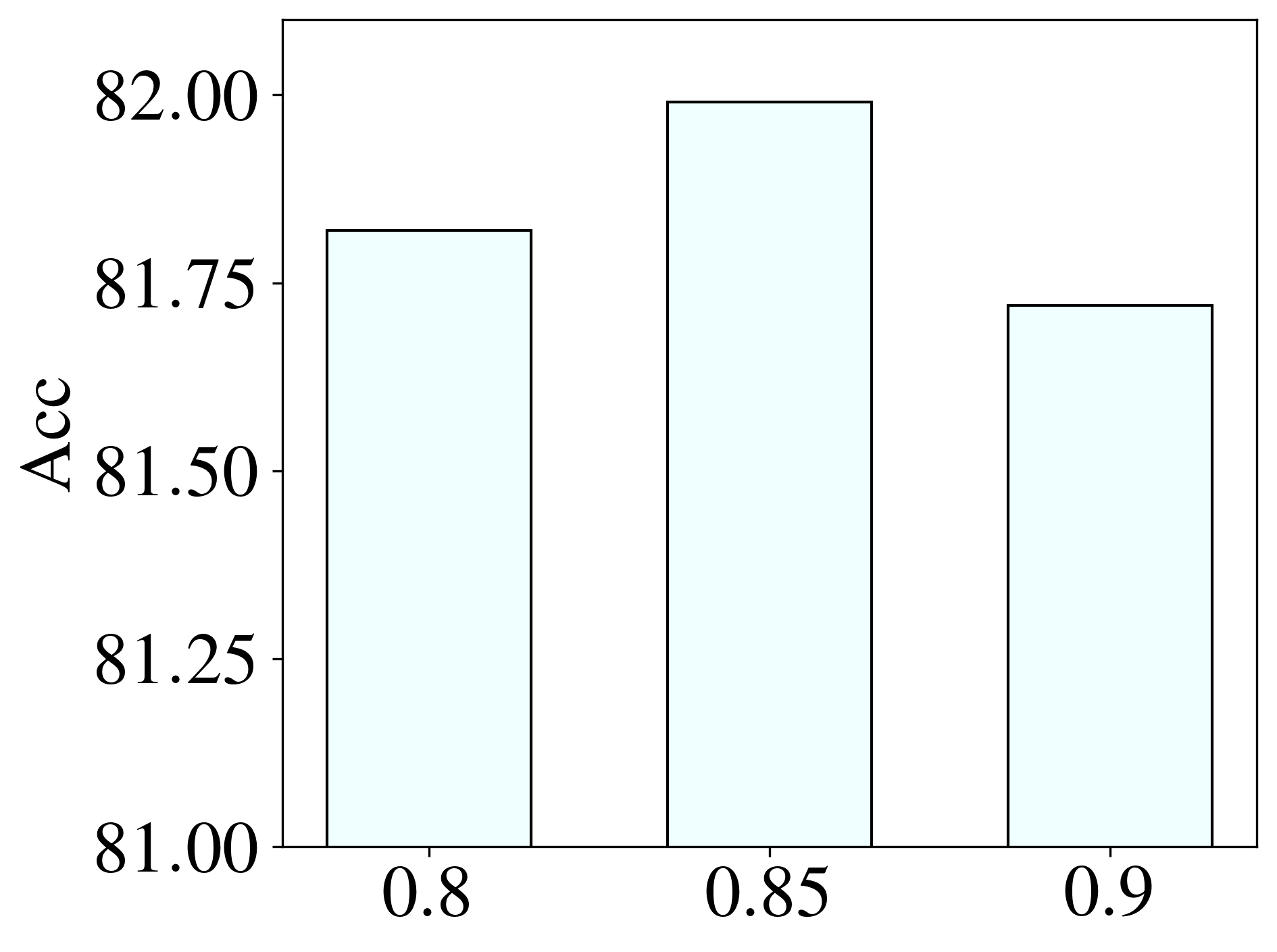}
		\vspace{-0.1in}
		\captionof{figure}{Results on MNLI-m under different exponential moving average parameters. Left: $\betaI$; Right: $\betaU$. By default, we set $\betaI=0.85$ and $\betaU=0.85$.}
		\label{tab:ablation-beta}
	\end{minipage} \hspace{0.1in}
	\begin{minipage}{0.35\textwidth}
		\vspace{-5mm}
		\centering
		\captionof{table}{\small Variants of the importance score. We prune a BERT-base model.}
		\vspace{-2mm}
		\label{tab:importance}
		\begin{tabular}{l|ccc}
			\toprule
			& {\bf SST-2} & {\bf RTE} & {\bf CoLA} \\ \midrule
			{\OurAlg} & 90.5 & 65.3 & 44.3 \\ \midrule
			$\Sc=\Ibar$ & 89.4 & 63.6 & 42.8 \\
			$\Sc=\Ubar$ & 89.3 & 64.3 & 40.6 \\
			$\Sc=\Ibar / \Ubar$ & 80.9 &  56.7 & \emph{N.A.}  \\
			\bottomrule
		\end{tabular}
	\end{minipage}
\end{table*}

\vspace{0.1in} \noindent
\textbf{Sensitivity to exponential moving average.}
Table~\ref{tab:ablation-beta} summarizes experimental results when we change the exponential moving average parameters $\betaI$ and $\betaU$.
In these experiments, we prune a pre-trained BERT-base model on the MNLI-m dataset.
From the results, we see that our method is robust to these two hyper-parameters.
In practice we fix $\betaI=\betaU=0.85$. Even when these selected hyper-parameters are not optimal, performance of {\OurAlg} is still better than the baseline methods.

\section{Extension to Structured Pruning}  
\label{sec:structured}

Algorithm~\ref{alg:our_algorithm} targets on unstructured pruning \citep{lecun1990optimal,han2015learning,frankle2018lottery,sanh2020movement}, which prunes model weights entry-wise.
However, it is notoriously difficult to speedup inference of the pruned model, which requires low-level implementation that is not supported by existing deep learning libraries such as PyTorch \citep{paszke2019pytorch} and Tensorflow \citep{abadi2016tensorflow}.
In contrast, structured pruning \citep{mccarley2019structured,fan2019reducing,sajjad2020poor,lagunas2021block} methods remove columns or blocks of the weights, such that the pruned model can be easily compressed. As such, inference speedup via structured pruning is more implementation-friendly.

\subsection{Uncertainty Adjusted Structured Pruning}
Our method can be extended to structured pruning. In this case, the proposed importance score is computed group-wise instead of entry-wise (c.f. \eqref{eq:importance_score}).
Specifically, we divided the parameter $\bTheta$ into $p$ disjoint groups denoted by 
\begin{align*}
    \bTheta = \{\bTheta_{\mathcal{G}_1}, \bTheta_{\mathcal{G}_2}, \dots, \bTheta_{\mathcal{G}_p}\}.
\end{align*}
For example, for a weight matrix, $\mathcal{G}_j$ can denote its $j$-th column.
Then the sensitivity of $ \bThetaGj $ is defined as
\begin{align}
    \I_{\Gj} = \left| \bThetaGj^{\top} \nabla_{\Gj} \cL(\bTheta) \right|. 
\end{align}
We apply exponential moving average to smooth the sensitivity of each group: 
\begin{align*}
    \Ibar^{(t+1)}_{\Gj} = \betaI\Ibart_{\Gj} + (1-\betaI) \It_{\Gj}. 
\end{align*}
Similar to \eqref{eq:uncertainty} and \eqref{eq:ema_uncertainty}, we quantify the uncertainty of sensitivity of each group and then apply exponential moving average as the following:
\begin{align*}
    & \Ut_{\Gj} = \left| \It_{\Gj} - \Ibart_{\Gj} \right|, \quad \Ubar^{(t+1)}_{\Gj} = \betaU \Ubart_{\Gj} + (1-\betaU) \Ut_{\Gj}. 
\end{align*}
The importance score is defined as (c.f. \eqref{eq:importance_score})
\begin{align*}
    &\Sct_{\Gj} = \Ibart_{\Gj} \cdot \Ubart_{\Gj}.
\end{align*}
We prune the model using \eqref{eq:proj_function} in a group fashion, i.e., we zero-out the entire group $\bThetaGj$ if $\Sct_{\Gj}$ is not in top-$\rr$\%. 

\begin{table}[h!]
	\vspace{-1mm}
	\caption{Results with $\text{BERT}_{\text{base}}$ on SQuAD v1.1. We report EM/F1. Here \textit{Ratio} is the proportion of remaining weights.}
	\vspace{-3mm}
	\label{tab:sturctured_experiemnts}
	\begin{center}
		\begin{tabular}{c|cc}
			\toprule
			{\bf Ratio} & {{\OurAlg}} & {{\OurAlg}\textsubscript{structure}}  \\
			\midrule 
			{40\%} & 78.0 / 86.9 & 75.6 / 84.9 \\
			{50\%} & 78.5 / 87.2 & 77.0 / 85.9 \\
			\bottomrule
		\end{tabular}
	\end{center}
	\vspace{-5mm}
\end{table}

\subsection{Experimental Results}
We evaluate performance of the proposed structured pruning method by pruning BERT-base on SQuAD v1.1.
Table~\ref{tab:sturctured_experiemnts} summarizes experimental results.
It is expected that performance of structured pruning is lower than the unstructured alternative, because we have less control over individual weights. For example, important weights and unimportant ones may locate in the same column and are pruned together.
From the results, we see that performance of {\OurAlg}\textsubscript{structure} drops by about $1.5$ and $2.0$ when the sparsity levels are $50\%$ and $60\%$, respectively.

\section{Discussion}\label{sec:discussion}
Movement pruning \citep{sanh2020movement} is one of the most popular model pruning methods. 
However, empirically we observe that it suffers from training instability and even divergence.
This might be due to the following reasons:

First, as mentioned earlier, the importance scores (sensitivity) estimated on mini-batches have high variability in movement pruning. This causes training instability as a weight may frequently alternates between being retained and being pruned. By considering uncertainty of the importance estimation, variability of importance scores in {\OurAlg} is drastically smaller (Figure~\ref{fig:sensitivity_distribution}, bottom), yielding a more stable training process and better model performance.

Second, a weight is masked instead of zeroed-out when it is deemed unimportant in movement pruning. Subsequently, if the rank of the weight's importance score raises in later iterations, it is unmasked and its value is restored.
However, the may be a number of iterations between a weight being masked and being reactivated, such that the weight's value becomes stale. In this case, such a stale weight can have a huge influence to the model, rendering training unstable.
In contrast, in {\OurAlg}, we let a weight starts from zero when reactivated.
This is natural because a pruned weight is still included in the training dynamics as a zero value.

\section{Conclusion}
\label{sec:conclusion}

We propose {\OurAlg}, which is a model pruning algorithm that considers both sensitivity of model weights and uncertainty of importance estimation.
In {\OurAlg}, we use exponential moving average to smooth the sensitivity computed on mini-batches. Moreover, we quantify uncertainty of the importance estimation via the local temporal variation. These approaches effectively reduce variability of the importance score (smoothed sensitivity times uncertainty) and stabilize training. 
We conduct extensive experiments on natural language understanding, question answering and image classification. Results show that {\OurAlg} significantly outperforms existing approaches. Our approach is particularly effective in the high sparsity regime.
For example, when pruning DeBERTaV3-base on the SQuAD v1.1 dataset, our method achieves a more than $5.0\%$ improvement when $90\%$ of the weights are pruned.
Moreover, training of {\OurAlg} is more stable and less sensitive to the hyper-parameters.



\bibliography{ref}

\begin{thebibliography}{55}
\expandafter\ifx\csname natexlab\endcsname\relax\def\natexlab#1{#1}\fi
\expandafter\ifx\csname url\endcsname\relax
  \def\url#1{\texttt{#1}}\fi
\expandafter\ifx\csname urlprefix\endcsname\relax\def\urlprefix{}\fi

\bibitem[{Abadi et~al.(2016)Abadi, Barham, Chen, Chen, Davis, Dean, Devin,
  Ghemawat, Irving, Isard et~al.}]{abadi2016tensorflow}
\textsc{Abadi, M.}, \textsc{Barham, P.}, \textsc{Chen, J.}, \textsc{Chen, Z.},
  \textsc{Davis, A.}, \textsc{Dean, J.}, \textsc{Devin, M.}, \textsc{Ghemawat,
  S.}, \textsc{Irving, G.}, \textsc{Isard, M.} \textsc{et~al.} (2016).
\newblock Tensorflow: A system for large-scale machine learning.
\newblock In \textit{12th $\{$USENIX$\}$ symposium on operating systems design
  and implementation ($\{$OSDI$\}$ 16)}.

\bibitem[{Bar-Haim et~al.(2006)Bar-Haim, Dagan, Dolan, Ferro and
  Giampiccolo}]{rte2}
\textsc{Bar-Haim, R.}, \textsc{Dagan, I.}, \textsc{Dolan, B.}, \textsc{Ferro,
  L.} and \textsc{Giampiccolo, D.} (2006).
\newblock The second {PASCAL} recognising textual entailment challenge.
\newblock In \textit{Proceedings of the Second {PASCAL} Challenges Workshop on
  Recognising Textual Entailment}.

\bibitem[{Behnke and Heafield(2021)}]{behnke2021pruning}
\textsc{Behnke, M.} and \textsc{Heafield, K.} (2021).
\newblock Pruning neural machine translation for speed using group lasso.
\newblock In \textit{Proceedings of the Sixth Conference on Machine
  Translation}.

\bibitem[{Bentivogli et~al.(2009)Bentivogli, Dagan, Dang, Giampiccolo and
  Magnini}]{rte5}
\textsc{Bentivogli, L.}, \textsc{Dagan, I.}, \textsc{Dang, H.~T.},
  \textsc{Giampiccolo, D.} and \textsc{Magnini, B.} (2009).
\newblock The fifth pascal recognizing textual entailment challenge.
\newblock In \textit{In Proc Text Analysis Conference (TAC’09}.

\bibitem[{Brown et~al.(2020)Brown, Mann, Ryder, Subbiah, Kaplan, Dhariwal,
  Neelakantan, Shyam, Sastry, Askell, Agarwal, Herbert{-}Voss, Krueger,
  Henighan, Child, Ramesh, Ziegler, Wu, Winter, Hesse, Chen, Sigler, Litwin,
  Gray, Chess, Clark, Berner, McCandlish, Radford, Sutskever and
  Amodei}]{brown2020language}
\textsc{Brown, T.~B.}, \textsc{Mann, B.}, \textsc{Ryder, N.}, \textsc{Subbiah,
  M.}, \textsc{Kaplan, J.}, \textsc{Dhariwal, P.}, \textsc{Neelakantan, A.},
  \textsc{Shyam, P.}, \textsc{Sastry, G.}, \textsc{Askell, A.},
  \textsc{Agarwal, S.}, \textsc{Herbert{-}Voss, A.}, \textsc{Krueger, G.},
  \textsc{Henighan, T.}, \textsc{Child, R.}, \textsc{Ramesh, A.},
  \textsc{Ziegler, D.~M.}, \textsc{Wu, J.}, \textsc{Winter, C.}, \textsc{Hesse,
  C.}, \textsc{Chen, M.}, \textsc{Sigler, E.}, \textsc{Litwin, M.},
  \textsc{Gray, S.}, \textsc{Chess, B.}, \textsc{Clark, J.}, \textsc{Berner,
  C.}, \textsc{McCandlish, S.}, \textsc{Radford, A.}, \textsc{Sutskever, I.}
  and \textsc{Amodei, D.} (2020).
\newblock Language models are few-shot learners.
\newblock In \textit{Advances in Neural Information Processing Systems 33:
  Annual Conference on Neural Information Processing Systems 2020, NeurIPS
  2020, December 6-12, 2020, virtual} (H.~Larochelle, M.~Ranzato, R.~Hadsell,
  M.~Balcan and H.~Lin, eds.).

\bibitem[{Cer et~al.(2017)Cer, Diab, Agirre, Lopez-Gazpio and
  Specia}]{sts-b2017}
\textsc{Cer, D.}, \textsc{Diab, M.}, \textsc{Agirre, E.}, \textsc{Lopez-Gazpio,
  I.} and \textsc{Specia, L.} (2017).
\newblock {S}em{E}val-2017 task 1: Semantic textual similarity multilingual and
  crosslingual focused evaluation.
\newblock In \textit{Proceedings of the 11th International Workshop on Semantic
  Evaluation ({S}em{E}val-2017)}. Association for Computational Linguistics,
  Vancouver, Canada.

\bibitem[{Chen et~al.(2020)Chen, Frankle, Chang, Liu, Zhang, Wang and
  Carbin}]{chen2020lottery}
\textsc{Chen, T.}, \textsc{Frankle, J.}, \textsc{Chang, S.}, \textsc{Liu, S.},
  \textsc{Zhang, Y.}, \textsc{Wang, Z.} and \textsc{Carbin, M.} (2020).
\newblock The lottery ticket hypothesis for pre-trained {BERT} networks.
\newblock In \textit{Advances in Neural Information Processing Systems 33:
  Annual Conference on Neural Information Processing Systems 2020, NeurIPS
  2020, December 6-12, 2020, virtual} (H.~Larochelle, M.~Ranzato, R.~Hadsell,
  M.~Balcan and H.~Lin, eds.).

\bibitem[{Dagan et~al.(2006)Dagan, Glickman and Magnini}]{rte1}
\textsc{Dagan, I.}, \textsc{Glickman, O.} and \textsc{Magnini, B.} (2006).
\newblock The pascal recognising textual entailment challenge.
\newblock In \textit{Proceedings of the First International Conference on
  Machine Learning Challenges: Evaluating Predictive Uncertainty Visual Object
  Classification, and Recognizing Textual Entailment}. MLCW'05,
  Springer-Verlag, Berlin, Heidelberg.

\bibitem[{Deng et~al.(2009)Deng, Dong, Socher, Li, Li and
  Li}]{deng2009imagenet}
\textsc{Deng, J.}, \textsc{Dong, W.}, \textsc{Socher, R.}, \textsc{Li, L.},
  \textsc{Li, K.} and \textsc{Li, F.} (2009).
\newblock Imagenet: {A} large-scale hierarchical image database.
\newblock In \textit{2009 {IEEE} Computer Society Conference on Computer Vision
  and Pattern Recognition {(CVPR} 2009), 20-25 June 2009, Miami, Florida,
  {USA}}. {IEEE} Computer Society.

\bibitem[{Devlin et~al.(2019)Devlin, Chang, Lee and Toutanova}]{devlin2018bert}
\textsc{Devlin, J.}, \textsc{Chang, M.-W.}, \textsc{Lee, K.} and
  \textsc{Toutanova, K.} (2019).
\newblock {BERT}: Pre-training of deep bidirectional transformers for language
  understanding.
\newblock In \textit{Proceedings of the 2019 Conference of the North {A}merican
  Chapter of the Association for Computational Linguistics: Human Language
  Technologies, Volume 1 (Long and Short Papers)}. Association for
  Computational Linguistics, Minneapolis, Minnesota.

\bibitem[{Ding et~al.(2019)Ding, Ding, Zhou, Guo, Han and Liu}]{ding2019global}
\textsc{Ding, X.}, \textsc{Ding, G.}, \textsc{Zhou, X.}, \textsc{Guo, Y.},
  \textsc{Han, J.} and \textsc{Liu, J.} (2019).
\newblock Global sparse momentum {SGD} for pruning very deep neural networks.
\newblock In \textit{Advances in Neural Information Processing Systems 32:
  Annual Conference on Neural Information Processing Systems 2019, NeurIPS
  2019, December 8-14, 2019, Vancouver, BC, Canada} (H.~M. Wallach,
  H.~Larochelle, A.~Beygelzimer, F.~d'Alch{\'{e}}{-}Buc, E.~B. Fox and
  R.~Garnett, eds.).

\bibitem[{Dolan and Brockett(2005)}]{mrpc2005}
\textsc{Dolan, W.~B.} and \textsc{Brockett, C.} (2005).
\newblock Automatically constructing a corpus of sentential paraphrases.
\newblock In \textit{Proceedings of the Third International Workshop on
  Paraphrasing ({IWP}2005)}.

\bibitem[{Dosovitskiy et~al.(2020)Dosovitskiy, Beyer, Kolesnikov, Weissenborn,
  Zhai, Unterthiner, Dehghani, Minderer, Heigold, Gelly
  et~al.}]{dosovitskiy2020image}
\textsc{Dosovitskiy, A.}, \textsc{Beyer, L.}, \textsc{Kolesnikov, A.},
  \textsc{Weissenborn, D.}, \textsc{Zhai, X.}, \textsc{Unterthiner, T.},
  \textsc{Dehghani, M.}, \textsc{Minderer, M.}, \textsc{Heigold, G.},
  \textsc{Gelly, S.} \textsc{et~al.} (2020).
\newblock An image is worth 16x16 words: Transformers for image recognition at
  scale.
\newblock \textit{arXiv preprint arXiv:2010.11929}.

\bibitem[{Fan et~al.(2020)Fan, Grave and Joulin}]{fan2019reducing}
\textsc{Fan, A.}, \textsc{Grave, E.} and \textsc{Joulin, A.} (2020).
\newblock Reducing transformer depth on demand with structured dropout.
\newblock In \textit{8th International Conference on Learning Representations,
  {ICLR} 2020, Addis Ababa, Ethiopia, April 26-30, 2020}. OpenReview.net.

\bibitem[{Frankle and Carbin(2019)}]{frankle2018lottery}
\textsc{Frankle, J.} and \textsc{Carbin, M.} (2019).
\newblock The lottery ticket hypothesis: Finding sparse, trainable neural
  networks.
\newblock In \textit{7th International Conference on Learning Representations,
  {ICLR} 2019, New Orleans, LA, USA, May 6-9, 2019}. OpenReview.net.

\bibitem[{Giampiccolo et~al.(2007)Giampiccolo, Magnini, Dagan and Dolan}]{rte3}
\textsc{Giampiccolo, D.}, \textsc{Magnini, B.}, \textsc{Dagan, I.} and
  \textsc{Dolan, B.} (2007).
\newblock The third {PASCAL} recognizing textual entailment challenge.
\newblock In \textit{Proceedings of the {ACL}-{PASCAL} Workshop on Textual
  Entailment and Paraphrasing}. Association for Computational Linguistics,
  Prague.

\bibitem[{Han et~al.(2016)Han, Liu, Mao, Pu, Pedram, Horowitz and
  Dally}]{han2016eie}
\textsc{Han, S.}, \textsc{Liu, X.}, \textsc{Mao, H.}, \textsc{Pu, J.},
  \textsc{Pedram, A.}, \textsc{Horowitz, M.~A.} and \textsc{Dally, W.~J.}
  (2016).
\newblock Eie: Efficient inference engine on compressed deep neural network.
\newblock \textit{ACM SIGARCH Computer Architecture News}, \textbf{44}
  243--254.

\bibitem[{Han et~al.(2015{\natexlab{a}})Han, Mao and Dally}]{han2015deep}
\textsc{Han, S.}, \textsc{Mao, H.} and \textsc{Dally, W.~J.}
  (2015{\natexlab{a}}).
\newblock Deep compression: Compressing deep neural networks with pruning,
  trained quantization and huffman coding.
\newblock \textit{arXiv preprint arXiv:1510.00149}.

\bibitem[{Han et~al.(2015{\natexlab{b}})Han, Pool, Tran and
  Dally}]{han2015learning}
\textsc{Han, S.}, \textsc{Pool, J.}, \textsc{Tran, J.} and \textsc{Dally,
  W.~J.} (2015{\natexlab{b}}).
\newblock Learning both weights and connections for efficient neural network.
\newblock In \textit{Advances in Neural Information Processing Systems 28:
  Annual Conference on Neural Information Processing Systems 2015, December
  7-12, 2015, Montreal, Quebec, Canada} (C.~Cortes, N.~D. Lawrence, D.~D. Lee,
  M.~Sugiyama and R.~Garnett, eds.).

\bibitem[{He et~al.(2021{\natexlab{a}})He, Gao and Chen}]{he2021debertav3}
\textsc{He, P.}, \textsc{Gao, J.} and \textsc{Chen, W.} (2021{\natexlab{a}}).
\newblock Debertav3: Improving deberta using electra-style pre-training with
  gradient-disentangled embedding sharing.
\newblock \textit{arXiv preprint arXiv:2111.09543}.

\bibitem[{He et~al.(2021{\natexlab{b}})He, Liu, Gao and Chen}]{he2021deberta}
\textsc{He, P.}, \textsc{Liu, X.}, \textsc{Gao, J.} and \textsc{Chen, W.}
  (2021{\natexlab{b}}).
\newblock Deberta: Decoding-enhanced bert with disentangled attention.
\newblock In \textit{International Conference on Learning Representations}.

\bibitem[{Krizhevsky(2009)}]{Krizhevsky09learningmultiple}
\textsc{Krizhevsky, A.} (2009).
\newblock Learning multiple layers of features from tiny images.
\newblock Tech. rep.

\bibitem[{Krizhevsky et~al.(2009)Krizhevsky, Hinton
  et~al.}]{krizhevsky2009learning}
\textsc{Krizhevsky, A.}, \textsc{Hinton, G.} \textsc{et~al.} (2009).
\newblock Learning multiple layers of features from tiny images.

\bibitem[{Lagunas et~al.(2021)Lagunas, Charlaix, Sanh and
  Rush}]{lagunas2021block}
\textsc{Lagunas, F.}, \textsc{Charlaix, E.}, \textsc{Sanh, V.} and
  \textsc{Rush, A.~M.} (2021).
\newblock Block pruning for faster transformers.
\newblock \textit{arXiv preprint arXiv:2109.04838}.

\bibitem[{Lai et~al.(1985)Lai, Robbins et~al.}]{lai1985asymptotically}
\textsc{Lai, T.~L.}, \textsc{Robbins, H.} \textsc{et~al.} (1985).
\newblock Asymptotically efficient adaptive allocation rules.
\newblock \textit{Advances in applied mathematics}, \textbf{6} 4--22.

\bibitem[{LeCun et~al.(1990)LeCun, Denker and Solla}]{lecun1990optimal}
\textsc{LeCun, Y.}, \textsc{Denker, J.~S.} and \textsc{Solla, S.~A.} (1990).
\newblock Optimal brain damage.
\newblock In \textit{Advances in neural information processing systems}.

\bibitem[{Lee et~al.(2019)Lee, Ajanthan and Torr}]{lee2018snip}
\textsc{Lee, N.}, \textsc{Ajanthan, T.} and \textsc{Torr, P. H.~S.} (2019).
\newblock Snip: single-shot network pruning based on connection sensitivity.
\newblock In \textit{7th International Conference on Learning Representations,
  {ICLR} 2019, New Orleans, LA, USA, May 6-9, 2019}. OpenReview.net.

\bibitem[{Levesque et~al.(2012)Levesque, Davis and Morgenstern}]{wnli2012}
\textsc{Levesque, H.}, \textsc{Davis, E.} and \textsc{Morgenstern, L.} (2012).
\newblock The winograd schema challenge.
\newblock In \textit{Thirteenth International Conference on the Principles of
  Knowledge Representation and Reasoning}.

\bibitem[{Liang et~al.(2021)Liang, Zuo, Chen, Jiang, Liu, He, Zhao and
  Chen}]{liang2021super}
\textsc{Liang, C.}, \textsc{Zuo, S.}, \textsc{Chen, M.}, \textsc{Jiang, H.},
  \textsc{Liu, X.}, \textsc{He, P.}, \textsc{Zhao, T.} and \textsc{Chen, W.}
  (2021).
\newblock Super tickets in pre-trained language models: From model compression
  to improving generalization.
\newblock In \textit{Proceedings of the 59th Annual Meeting of the Association
  for Computational Linguistics and the 11th International Joint Conference on
  Natural Language Processing (Volume 1: Long Papers)}. Association for
  Computational Linguistics, Online.

\bibitem[{Liu et~al.(2019{\natexlab{a}})Liu, He, Chen and Gao}]{liu2019multi}
\textsc{Liu, X.}, \textsc{He, P.}, \textsc{Chen, W.} and \textsc{Gao, J.}
  (2019{\natexlab{a}}).
\newblock Multi-task deep neural networks for natural language understanding.
\newblock In \textit{Proceedings of the 57th Annual Meeting of the Association
  for Computational Linguistics}. Association for Computational Linguistics,
  Florence, Italy.

\bibitem[{Liu et~al.(2020)Liu, Wang, Ji, Cheng, Zhu, Awa, He, Chen, Poon, Cao
  and Gao}]{mtdnn2020demo}
\textsc{Liu, X.}, \textsc{Wang, Y.}, \textsc{Ji, J.}, \textsc{Cheng, H.},
  \textsc{Zhu, X.}, \textsc{Awa, E.}, \textsc{He, P.}, \textsc{Chen, W.},
  \textsc{Poon, H.}, \textsc{Cao, G.} and \textsc{Gao, J.} (2020).
\newblock The {M}icrosoft toolkit of multi-task deep neural networks for
  natural language understanding.
\newblock In \textit{Proceedings of the 58th Annual Meeting of the Association
  for Computational Linguistics: System Demonstrations}. Association for
  Computational Linguistics, Online.

\bibitem[{Liu et~al.(2019{\natexlab{b}})Liu, Ott, Goyal, Du, Joshi, Chen, Levy,
  Lewis, Zettlemoyer and Stoyanov}]{liu2019roberta}
\textsc{Liu, Y.}, \textsc{Ott, M.}, \textsc{Goyal, N.}, \textsc{Du, J.},
  \textsc{Joshi, M.}, \textsc{Chen, D.}, \textsc{Levy, O.}, \textsc{Lewis, M.},
  \textsc{Zettlemoyer, L.} and \textsc{Stoyanov, V.} (2019{\natexlab{b}}).
\newblock Roberta: A robustly optimized bert pretraining approach.
\newblock \textit{arXiv preprint arXiv:1907.11692}.

\bibitem[{Loshchilov and Hutter(2019)}]{loshchilov2017decoupled}
\textsc{Loshchilov, I.} and \textsc{Hutter, F.} (2019).
\newblock Decoupled weight decay regularization.
\newblock In \textit{7th International Conference on Learning Representations,
  {ICLR} 2019, New Orleans, LA, USA, May 6-9, 2019}. OpenReview.net.

\bibitem[{Louizos et~al.(2018)Louizos, Welling and
  Kingma}]{louizos2017learning}
\textsc{Louizos, C.}, \textsc{Welling, M.} and \textsc{Kingma, D.~P.} (2018).
\newblock Learning sparse neural networks through l{\_}0 regularization.
\newblock In \textit{6th International Conference on Learning Representations,
  {ICLR} 2018, Vancouver, BC, Canada, April 30 - May 3, 2018, Conference Track
  Proceedings}. OpenReview.net.

\bibitem[{Mallya and Lazebnik(2018)}]{mallya2018piggyback}
\textsc{Mallya, A.} and \textsc{Lazebnik, S.} (2018).
\newblock Piggyback: Adding multiple tasks to a single, fixed network by
  learning to mask.
\newblock \textit{arXiv preprint arXiv:1801.06519}, \textbf{6}.

\bibitem[{McCarley et~al.(2019)McCarley, Chakravarti and
  Sil}]{mccarley2019structured}
\textsc{McCarley, J.}, \textsc{Chakravarti, R.} and \textsc{Sil, A.} (2019).
\newblock Structured pruning of a bert-based question answering model.
\newblock \textit{arXiv preprint arXiv:1910.06360}.

\bibitem[{Molchanov et~al.(2019)Molchanov, Mallya, Tyree, Frosio and
  Kautz}]{molchanov2019importance}
\textsc{Molchanov, P.}, \textsc{Mallya, A.}, \textsc{Tyree, S.},
  \textsc{Frosio, I.} and \textsc{Kautz, J.} (2019).
\newblock Importance estimation for neural network pruning.
\newblock In \textit{{IEEE} Conference on Computer Vision and Pattern
  Recognition, {CVPR} 2019, Long Beach, CA, USA, June 16-20, 2019}. Computer
  Vision Foundation / {IEEE}.

\bibitem[{Molchanov et~al.(2017)Molchanov, Tyree, Karras, Aila and
  Kautz}]{molchanov2016pruning}
\textsc{Molchanov, P.}, \textsc{Tyree, S.}, \textsc{Karras, T.}, \textsc{Aila,
  T.} and \textsc{Kautz, J.} (2017).
\newblock Pruning convolutional neural networks for resource efficient
  inference.
\newblock In \textit{5th International Conference on Learning Representations,
  {ICLR} 2017, Toulon, France, April 24-26, 2017, Conference Track
  Proceedings}. OpenReview.net.

\bibitem[{Paganini and Forde(2020)}]{paganini2020iterative}
\textsc{Paganini, M.} and \textsc{Forde, J.} (2020).
\newblock On iterative neural network pruning, reinitialization, and the
  similarity of masks.
\newblock \textit{arXiv preprint arXiv:2001.05050}.

\bibitem[{Paszke et~al.(2019)Paszke, Gross, Massa, Lerer, Bradbury, Chanan,
  Killeen, Lin, Gimelshein, Antiga, Desmaison, K{\"{o}}pf, Yang, DeVito,
  Raison, Tejani, Chilamkurthy, Steiner, Fang, Bai and
  Chintala}]{paszke2019pytorch}
\textsc{Paszke, A.}, \textsc{Gross, S.}, \textsc{Massa, F.}, \textsc{Lerer,
  A.}, \textsc{Bradbury, J.}, \textsc{Chanan, G.}, \textsc{Killeen, T.},
  \textsc{Lin, Z.}, \textsc{Gimelshein, N.}, \textsc{Antiga, L.},
  \textsc{Desmaison, A.}, \textsc{K{\"{o}}pf, A.}, \textsc{Yang, E.},
  \textsc{DeVito, Z.}, \textsc{Raison, M.}, \textsc{Tejani, A.},
  \textsc{Chilamkurthy, S.}, \textsc{Steiner, B.}, \textsc{Fang, L.},
  \textsc{Bai, J.} and \textsc{Chintala, S.} (2019).
\newblock Pytorch: An imperative style, high-performance deep learning library.
\newblock In \textit{Advances in Neural Information Processing Systems 32:
  Annual Conference on Neural Information Processing Systems 2019, NeurIPS
  2019, December 8-14, 2019, Vancouver, BC, Canada} (H.~M. Wallach,
  H.~Larochelle, A.~Beygelzimer, F.~d'Alch{\'{e}}{-}Buc, E.~B. Fox and
  R.~Garnett, eds.).

\bibitem[{Qian(1999)}]{qian1999momentum}
\textsc{Qian, N.} (1999).
\newblock On the momentum term in gradient descent learning algorithms.
\newblock \textit{Neural networks}, \textbf{12} 145--151.

\bibitem[{Radford et~al.(2019)Radford, Wu, Child, Luan, Amodei, Sutskever
  et~al.}]{radford2019language}
\textsc{Radford, A.}, \textsc{Wu, J.}, \textsc{Child, R.}, \textsc{Luan, D.},
  \textsc{Amodei, D.}, \textsc{Sutskever, I.} \textsc{et~al.} (2019).
\newblock Language models are unsupervised multitask learners.
\newblock \textit{OpenAI blog}, \textbf{1} 9.

\bibitem[{Rajpurkar et~al.(2016{\natexlab{a}})Rajpurkar, Zhang, Lopyrev and
  Liang}]{squad1}
\textsc{Rajpurkar, P.}, \textsc{Zhang, J.}, \textsc{Lopyrev, K.} and
  \textsc{Liang, P.} (2016{\natexlab{a}}).
\newblock {SQ}u{AD}: 100,000+ questions for machine comprehension of text.
\newblock In \textit{Proceedings of the 2016 Conference on Empirical Methods in
  Natural Language Processing}. Association for Computational Linguistics,
  Austin, Texas.

\bibitem[{Rajpurkar et~al.(2016{\natexlab{b}})Rajpurkar, Zhang, Lopyrev and
  Liang}]{rajpurkar2016squad}
\textsc{Rajpurkar, P.}, \textsc{Zhang, J.}, \textsc{Lopyrev, K.} and
  \textsc{Liang, P.} (2016{\natexlab{b}}).
\newblock {SQ}u{AD}: 100,000+ questions for machine comprehension of text.
\newblock In \textit{Proceedings of the 2016 Conference on Empirical Methods in
  Natural Language Processing}. Association for Computational Linguistics,
  Austin, Texas.

\bibitem[{Renda et~al.(2020)Renda, Frankle and Carbin}]{renda2020comparing}
\textsc{Renda, A.}, \textsc{Frankle, J.} and \textsc{Carbin, M.} (2020).
\newblock Comparing rewinding and fine-tuning in neural network pruning.
\newblock In \textit{8th International Conference on Learning Representations,
  {ICLR} 2020, Addis Ababa, Ethiopia, April 26-30, 2020}. OpenReview.net.

\bibitem[{Sajjad et~al.(2020)Sajjad, Dalvi, Durrani and Nakov}]{sajjad2020poor}
\textsc{Sajjad, H.}, \textsc{Dalvi, F.}, \textsc{Durrani, N.} and
  \textsc{Nakov, P.} (2020).
\newblock Poor man's bert: Smaller and faster transformer models.
\newblock \textit{arXiv e-prints} arXiv--2004.

\bibitem[{Sanh et~al.(2020)Sanh, Wolf and Rush}]{sanh2020movement}
\textsc{Sanh, V.}, \textsc{Wolf, T.} and \textsc{Rush, A.~M.} (2020).
\newblock Movement pruning: Adaptive sparsity by fine-tuning.

\bibitem[{Socher et~al.(2013)Socher, Perelygin, Wu, Chuang, Manning, Ng and
  Potts}]{sst2013}
\textsc{Socher, R.}, \textsc{Perelygin, A.}, \textsc{Wu, J.}, \textsc{Chuang,
  J.}, \textsc{Manning, C.~D.}, \textsc{Ng, A.} and \textsc{Potts, C.} (2013).
\newblock Recursive deep models for semantic compositionality over a sentiment
  treebank.
\newblock In \textit{Proceedings of the 2013 Conference on Empirical Methods in
  Natural Language Processing}. Association for Computational Linguistics,
  Seattle, Washington, USA.

\bibitem[{Wang et~al.(2019)Wang, Singh, Michael, Hill, Levy and
  Bowman}]{wang2018glue}
\textsc{Wang, A.}, \textsc{Singh, A.}, \textsc{Michael, J.}, \textsc{Hill, F.},
  \textsc{Levy, O.} and \textsc{Bowman, S.~R.} (2019).
\newblock {GLUE:} {A} multi-task benchmark and analysis platform for natural
  language understanding.
\newblock In \textit{7th International Conference on Learning Representations,
  {ICLR} 2019, New Orleans, LA, USA, May 6-9, 2019}. OpenReview.net.

\bibitem[{Warstadt et~al.(2019)Warstadt, Singh and Bowman}]{cola2018}
\textsc{Warstadt, A.}, \textsc{Singh, A.} and \textsc{Bowman, S.~R.} (2019).
\newblock Neural network acceptability judgments.
\newblock \textit{Transactions of the Association for Computational
  Linguistics}, \textbf{7} 625--641.

\bibitem[{Williams et~al.(2018)Williams, Nangia and Bowman}]{mnli2018}
\textsc{Williams, A.}, \textsc{Nangia, N.} and \textsc{Bowman, S.} (2018).
\newblock A broad-coverage challenge corpus for sentence understanding through
  inference.
\newblock In \textit{Proceedings of the 2018 Conference of the North {A}merican
  Chapter of the Association for Computational Linguistics: Human Language
  Technologies, Volume 1 (Long Papers)}. Association for Computational
  Linguistics, New Orleans, Louisiana.

\bibitem[{Wolf et~al.(2019)Wolf, Debut, Sanh, Chaumond, Delangue, Moi, Cistac,
  Rault, Louf, Funtowicz et~al.}]{wolf2019huggingface}
\textsc{Wolf, T.}, \textsc{Debut, L.}, \textsc{Sanh, V.}, \textsc{Chaumond,
  J.}, \textsc{Delangue, C.}, \textsc{Moi, A.}, \textsc{Cistac, P.},
  \textsc{Rault, T.}, \textsc{Louf, R.}, \textsc{Funtowicz, M.} \textsc{et~al.}
  (2019).
\newblock Huggingface's transformers: State-of-the-art natural language
  processing.
\newblock \textit{ArXiv preprint}, \textbf{abs/1910.03771}.

\bibitem[{Zafrir et~al.(2021)Zafrir, Larey, Boudoukh, Shen and
  Wasserblat}]{zafrir2021prune}
\textsc{Zafrir, O.}, \textsc{Larey, A.}, \textsc{Boudoukh, G.}, \textsc{Shen,
  H.} and \textsc{Wasserblat, M.} (2021).
\newblock Prune once for all: Sparse pre-trained language models.
\newblock \textit{Advances in Neural Information Processing Systems}.

\bibitem[{Zhang et~al.(2021)Zhang, Wipf, Gan and Song}]{zhang2021biased}
\textsc{Zhang, Q.}, \textsc{Wipf, D.}, \textsc{Gan, Q.} and \textsc{Song, L.}
  (2021).
\newblock A biased graph neural network sampler with near-optimal regret.
\newblock \textit{Advances in Neural Information Processing Systems},
  \textbf{34} 8833--8844.

\bibitem[{Zhu and Gupta(2018)}]{zhu2017prune}
\textsc{Zhu, M.} and \textsc{Gupta, S.} (2018).
\newblock To prune, or not to prune: Exploring the efficacy of pruning for
  model compression.

\end{thebibliography}
\bibliographystyle{ims}

\newpage
\appendix
\onecolumn

\section{Sparsity Ratio Schedule}
\label{app:schedule}

The cubic schedule of sparsity ratio is widely applied by many existing methods \citep{zhu2017prune,sanh2020movement,zafrir2021prune}, which includes initial and final warmups: 
\begin{equation}\label{eq:cubic_schedule}
\rt = \begin{cases}
\rr^{(0)} & 0 \leq t<t_{i} \\ 
\rr^{(T)}+ \left(\rr^{(0)}-\rr^{(T)}\right)\left(1-\frac{t-t_{i}-t_{f}}{T-t_i - t_f}\right)^{3} & t_{i} \leq t<T-t_{f} \\ 
\rr^{(T)} &\text { o.w. }
\end{cases}
\end{equation}
Same as \citet{sanh2020movement}, we applies this cubic schedule to all baseline methods and our algorithm.

\section{GLUE Dataset Statistics}
\label{app:dataset-glue}

We present the dataset statistics of GLUE \citep{wang2018glue} in the following table. 
\begin{table*}[htb!]
	\begin{center}
		\begin{tabular}{l|l|c|c|c|c|c}
			\toprule 
			\bf Corpus &Task& \#Train & \#Dev & \#Test   & \#Label &Metrics\\ \midrule
			\multicolumn{6}{@{\hskip1pt}r@{\hskip1pt}}{Single-Sentence Classification (GLUE)} \\ \hline
			CoLA & Acceptability&8.5k & 1k & 1k & 2 & Matthews corr\\ \hline
			SST & Sentiment&67k & 872 & 1.8k & 2 & Accuracy\\ \midrule
			\multicolumn{6}{@{\hskip1pt}r@{\hskip1pt}}{Pairwise Text Classification (GLUE)} \\ \hline
			MNLI & NLI& 393k& 20k & 20k& 3 & Accuracy\\ \hline
			RTE & NLI &2.5k & 276 & 3k & 2 & Accuracy \\ \hline
			QQP & Paraphrase&364k & 40k & 391k& 2 & Accuracy/F1\\ \hline
			MRPC & Paraphrase &3.7k & 408 & 1.7k& 2&Accuracy/F1\\ \hline
			QNLI & QA/NLI& 108k &5.7k&5.7k&2& Accuracy\\ \midrule
			\multicolumn{5}{@{\hskip1pt}r@{\hskip1pt}}{Text Similarity (GLUE)} \\ \hline
			STS-B & Similarity &7k &1.5k& 1.4k &1 & Pearson/Spearman corr\\ \bottomrule
		\end{tabular}
	\end{center}
	\vskip -0.05in
	\caption{Summary of the GLUE benchmark.}
	\label{tab:glue}
\end{table*}

\section{Natural Language Understanding}\label{sec:app_nlu}

\subsection{Training Details} 

{\bf Implementation Details.} The implementation of {\OurAlg} on BERT-base is based on publicly available \textit{MT-DNN} \citep{liu2019multi,mtdnn2020demo}\footnote{\url{https://github.com/microsoft/MT-DNN}} code-base. The implementation of DeBertaV3-base \citep{he2021debertav3} on GLUE is based on \textit{Huggingface Transformers}\footnote{\url{https://github.com/huggingface/transformers}} \citep{wolf2019huggingface} code-base.

{\bf Hyper-parameter Details.} We select $ \betaI $ in range of $ \{0.75, 0.80, 0.85, 0.90\} $, find $ 0.85 $ generally perform best and fix it as $ 0.85 $ for all experiments. We select $ \betaU $ from $ \{0.850, 0.900, 0.950, 0.975\} $. We choose AdamW as optimizer and select learning rate from $\{3\times 10^{-5}, 5\times 10^{-5}, 8\times 10^{-5}, 1\times 10^{-4}  \}$ and batch size from $\{ 8, 16, 32 \}$ and fix them for each dataset. Table~\ref{tab:app_glue_setup} lists the detailed setup of each dataset. For baseline methods, we set the hyper-parameters all as reported by \citep{sanh2020movement}.

We found that {\OurAlg} is not sensitive to its hyper-parameters $ \betaI $ and $ \betaU $. The performance of {\OurAlg} does not alter dramatically among different hyper-parameter setup. 

\begin{table*}[h!]
\vspace{-1mm}
\caption{Hyper-parameter setup of {\OurAlg} for GLUE benchmark.}
\label{tab:app_glue_setup}
\begin{center}
\begin{small}
\begin{tabular}{l|c|cccccccc}
\toprule
\multirow{1}*{\bf Ratio} & {\bf Hyper-parameter} & {\bf MNLI} & {\bf RTE} & {\bf QNLI}  & {\bf MRPC} & {\bf QQP } & {\bf SST-2} & {\bf CoLA} & {\bf STS-B} \\ 
\midrule 
~ & \# epochs &   8  &  20 & 10 & 10 & 10 & 6 & 15 & 15 \\
~ & Batch size &   32 &  16 & 32 & 8 & 32 & 32 & 32 & 16 \\
~ & Learning rate & $ 8\times 10^{-5}$ & $1\times 10^{-4}$ & $ 1\times 10^{-4} $ & $ 1\times 10^{-4} $  & $ 1\times 10^{-4} $ & $ 8\times 10^{-5}$ & $ 1\times 10^{-4} $ & $1\times 10^{-4}$ \\
\midrule
\multirow{5}*{20\%} & $ t_i $ & 5400 & 200 & 2000 & 300 & 5400 & 1000 & 500 & 500 \\
~ & $ t_f $ &  22000 & 1200 & 12000 & 900 & 22000 & 5000 & 1500 & 2500 \\
~ & $ \betaI $ & 0.85 & 0.85 & 0.85 & 0.85 & 0.85 & 0.85 & 0.85 & 0.85 \\
~ & $ \betaU $ &  0.85 & 0.99 & 0.95 & 0.95 & 0.85 & 0.85 & 0.95 & 0.85  \\
\midrule 
\multirow{5}*{15\%} & $ t_i $ & 5400 & 200 & 2000 & 300 & 5400 & 1000 & 500 & 500  \\
~ & $ t_f $ &  22000 & 1200 & 12000 & 900 & 22000 & 5000 & 1500 & 2500 \\
~ & $ \betaI $ &  0.85 & 0.85 & 0.85 & 0.85 & 0.85 & 0.85 & 0.85 & 0.85 \\
~ & $ \betaU $ &  0.90 & 0.50 & 0.90 &0.95 & 0.90 & 0.90 & 0.975 & 0.90 \\
\midrule 
\multirow{5}*{10\%} & $ t_i $ & 5400 & 200 & 2000 & 300 & 5400 & 1000 & 500 & 500  \\
~ & $ t_f $ &  22000 & 1200 & 12000 & 900 & 22000 & 5000 & 1500 & 2500 \\
~ & $ \betaI $ & 0.85 & 0.85 & 0.85 & 0.85 & 0.85 & 0.85 & 0.85 & 0.85 \\
~ & $ \betaU $ & 0.85 & 0.95 & 0.95 & 0.95 & 0.90 & 0.85 & 0.95 & 0.95 \\
\bottomrule
\end{tabular}
\end{small}
\end{center}
\end{table*}

\section{Question Answering}\label{sec:app_squad}

\subsection{Dataset}
Following \citet{sanh2020movement}, we also choose SQuAD v1.1 \citep{rajpurkar2016squad} to evaluate the performance of {\OurAlg} on question answering task.

\subsection{Training Details}
We set the batch size as 16, the number of epochs for fine-tuning as 10, the optimizer as AdamW and the learning rate as $ 3\times 10^{-5} $ for all algorithms. We select the number of epochs as 10. 
For {\OurAlg}, we set its cubic sparsity schedule as $ t_i = 5400 $ and $ t_f = 22000 $. The baselines are all configured as reported by \citet{sanh2020movement}. The other hyper-parameters are reported in Table~\ref{tab:app_squad_setup}.

\begin{table*}[htb!]
\vspace{-2mm}
\caption{Hyper-parameter setup of {\OurAlg} on question answering tasks (SQuAD v1.1, \citet{rajpurkar2016squad}).}
\vspace{3mm}
\label{tab:app_squad_setup}
\begin{center}
\begin{small}
\begin{tabular}{l|cccccc}
\toprule
\multirow{1}*{\bf Ratio} & {10\%}  & {15\%}  & {20\%}  & {30\%} & {40\%} & {50\%}  \\ 
\midrule 
$ \betaI $ & 0.85 &  0.85  & 0.85  & 0.85  & 0.85  & 0.85 \\
$ \betaU $ & 0.975 & 0.975 & 0.975 & 0.975 & 0.900 & 0.975 \\
\bottomrule
\end{tabular}
\end{small}
\end{center}
\end{table*}

\section{Image Classification}\label{sec:app_vit}
For all image classification tasks we use the Vision Transformer model (ViT) \citep{dosovitskiy2020image}. We use the base model with an input patch size of $16 \times 16$. The ViT model is pre-trained on the Imagenet-21K dataset.

\subsection{Datasets}
We evaluate the {\OurAlg} method on two datasets, CIFAR-100 and the ILSVRC-2012 ImageNet dataset. The CIFAR-100 has 100 classes containing 600 images each. We use a image resolution of $224 \time 224$ for CIFAR-100. There are 500 training images and 100 testing images per class. The ILSVRC-2012 ImageNet dataset contains 1.2 million images spanning 1000 categories. We use a image resolution of $384 \times 384$ for ImageNet.

\subsection{Training Details}
{\bf Implementation Details.} The implementation of ViT follows the codebase of its pytorch version\footnote{\url{https://github.com/jeonsworld/ViT-pytorch}}. We use SGD with momentum \citep{qian1999momentum} as optimizer. For CIFAR100, we set batch size as 512 and learning rate as $0.03$. For ImageNet, we set batch size as 150 and learning rate as 0.03.

\textbf{Finetuning Details.} We fine-tune the ViT model with a learning rate 0f 0.03. We use a batch size of 512 for CIFAR-100 and 150 for ImageNet due to memory constraints. In addition, we use gradient clipping at norm 1. We also use a cosine decay for the learning rate with a warmup of 1000 steps.

\textbf{{\OurAlg} Details.}
We use the following hyperparameters for the {\OurAlg} algorithm. For the ImageNet dataset we use a $t_i = 7000$ step warmup period, a $t_f = 10000$ step final warmup period, and finetune for 60000 steps total. We set the exponential moving average parameter to 0.85.

For the CIFAR-100 dataset we use a 2000 step warmup period, a 8000 step final warmup period, and finetune for 20000 steps total. We set the exponential moving average parameter to 0.85.

\end{document}